\documentclass{article}

\usepackage{microtype}
\usepackage{graphicx}
\usepackage{booktabs}
\usepackage{hyperref}

\usepackage[preprint]{icml2026}

\usepackage{times}
\usepackage[T1]{fontenc}
\usepackage[utf8]{inputenc}
\usepackage{amsmath,amsfonts,amssymb}
\usepackage{algorithm}
\usepackage{algpseudocode}
\usepackage{xspace}
\usepackage{pifont}
\usepackage{xcolor}
\definecolor{darkgreen}{rgb}{0.0,0.5,0.0}
\usepackage{multirow}

\graphicspath{{./figures/}}

\newcommand{\methodname}{\textsc{Reflect}\xspace}
\newcommand{\tss}[1]{{\tiny\,$\pm$#1}}
\newcommand{\halfmark}{\ding{72}}

\icmltitlerunning{REFLECT: Intervention-Supported Error Attribution for Silent Failures}

\begin{document}

\twocolumn[
  \icmltitle{REFLECT: Intervention-Supported Error Attribution for Silent Failures in LLM Agent Traces}

  \begin{icmlauthorlist}
    \icmlauthor{Xiaofeng Lin}{ucla}
    \icmlauthor{Yingxu Wang}{ucla}
    \icmlauthor{Tung Sum Thomas Kwok}{ucla}
    \icmlauthor{Daniel Guo}{ucla}
    \icmlauthor{Sahil Arun Nale}{ucla}
    \icmlauthor{Charles Fleming}{cisco,um}
    \icmlauthor{Guang Cheng}{ucla}
  \end{icmlauthorlist}

  \icmlaffiliation{ucla}{University of California, Los Angeles, USA}
  \icmlaffiliation{cisco}{Cisco, USA}
  \icmlaffiliation{um}{University of Mississippi, USA}

  \icmlcorrespondingauthor{Guang Cheng}{guangcheng@stat.ucla.edu}

  \icmlkeywords{LLM agents, error localization, silent failures, trace analysis}

  \vskip 0.3in
]

\printAffiliationsAndNotice{Accepted at the ICML 2026 Workshop on Failure Modes in Agentic AI (FAGEN).}

\begin{abstract}
Large language model (LLM) agents now solve complex tasks through long plan-and-execution traces, yet the ability to locate errors in a completed trace still lags far behind, especially in the \emph{silent failure} regime.
Existing approaches predict suspect steps via classifiers or LLM judges, or recover correct answers via retry, but none feed the intervention outcome back to \emph{refine the attribution itself}.
We propose \methodname, a method that closes this gap by diagnosing a candidate error step, testing it through controlled replay with a diagnosis-specific patch, and using the verified outcome flip as contrastive evidence to refine the final attribution.
On four localization benchmarks spanning multiple domains, \methodname achieves the highest localization accuracy among same-auditor methods on all four benchmarks, with the largest gains on structured tool-use traces, while providing actionable localization even when ground-truth answers are unavailable.
\end{abstract}
\section{Introduction}
\label{sec:intro}

\begin{figure}[t]
\centering
\includegraphics[width=\columnwidth]{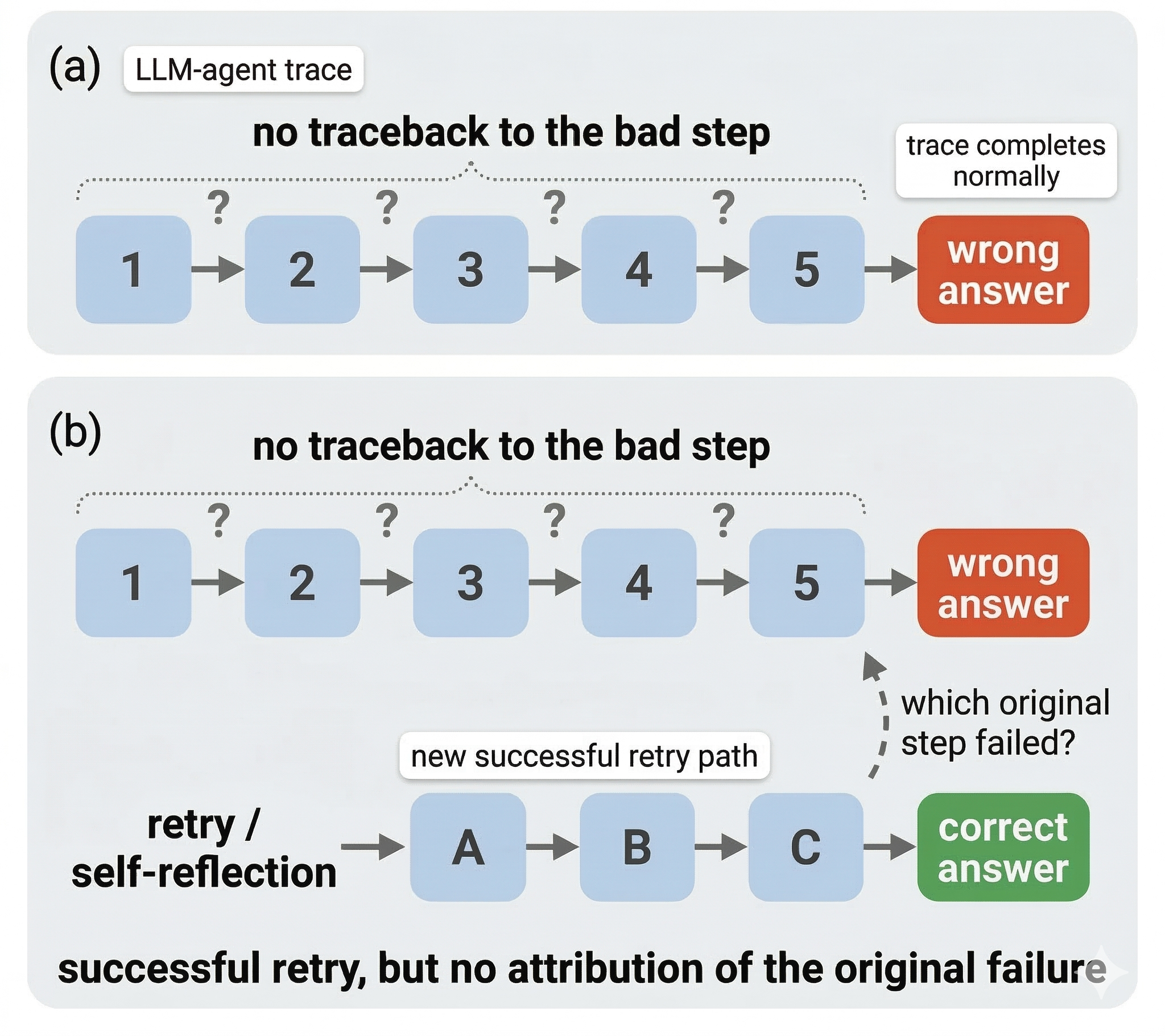}
\caption{Silent failure and the correction–attribution gap. (a) The trace completes normally but produces a wrong answer; no step signals an error. (b) Retry or self-reflection may recover the correct answer via an independent path, but this success does not identify which original step was decisive.}
\label{fig:correction-vs-attribution}
\end{figure}

While large language model (LLM) agents can now solve very complex tasks through \emph{long plan-and-execution traces}~\cite{yao2023react,yao2023tree,guo2024multiagents,plaat2025agentic,plaat2024reasoning}, their ability to \emph{review a completed trace and reliably localize where it went wrong} still lags far behind~\cite{huang2023selfcorrect,tyen2024mistakefinding,srivatsa2025cannot_spot_math_errors}.
That gap matters directly for deployment~\cite{jaffe2024real_world_workplaces,gartner2025autonomous_agents_survey,yehudai2025survey_evaluation_agents,yu2025trustworthy_agents_survey,rombaut2025watson,googleSREpostmortem}: AI cannot reliably boost productivity if humans must still step in to inspect completed work and find the source of failure, and agents cannot be trusted on high-stakes tasks unless they can localize errors rather than merely output plausible answers.

This challenge is especially acute in the \emph{silent failure} regime, where state-of-the-art agents complete without runtime exceptions (no tool crashes, no malformed JSON, no obvious runtime signal) and yet the final answer is still wrong. As agents grow more capable, silent failures become the dominant mode: the outputs look well-formed, and only the semantics are wrong~\cite{turpin2023cotunfaithful,azaria2023internal,liang2024internal,sun2024toolsfail}. While techniques based on reflection, retry, or stronger reasoning~\cite{wang2022selfconsistency,shinn2023reflexion,chen2023selfdebug,madaan2023selfrefine,openai2024learning_reason_with_llms,guo2025deepseekr1} can increase task accuracy, they do not resolve this problem, because \textbf{correction is not attribution}.
A system can recover the correct answer by resampling or generating a different trajectory, yet still fail to identify which step in the \emph{original} trace caused the failure. 

Concurrent work has explored locating errors in complex agent traces, but existing approaches exhibit three recurring gaps: \emph{localization without execution grounding}: trained classifiers~\cite{agentracer2025localization,liu2026trajad} and LLM judges~\cite{trail2025tracelocalization} predict suspect steps but cannot verify those predictions, and are known to hallucinate confident but wrong attributions~\cite{huang2023selfcorrect,tyen2024mistakefinding}; \emph{correction without attribution coupling}: retry and backtracking methods~\cite{ics2025backtrack,shinn2023reflexion} may recover correct answers but do not identify which step in the original trace was responsible, treating correction as an end rather than evidence; and \emph{signal checking without causal testing}: constraint and anomaly detectors~\cite{agentrx2026,thinkprm2025,lightman2023letsverify,vacareanu2024generalpurposeverification} flag suspicious steps but do not verify whether fixing the flagged step actually changes the outcome. These gaps share a common root: \textbf{attribution hypotheses are stated but not tested.}

We therefore identify four requirements for faithful error attribution in agent traces, which together distinguish \emph{tested attribution} from \emph{untested prediction}: (1) attribution must be grounded in execution outcomes, not narrative explanation; (2) evidence must derive from the same trace via prefix-preserving replay, not from independent retries; (3) replay must be targeted: constrained by a diagnosis-specific intervention, rather than bare resampling; and (4) attribution must be computed at inference time on the test trace, not predicted by a pre-trained classifier. Existing methods satisfy at most two of these requirements (Table~\ref{tab:method_comparison}).

To instantiate these requirements, we propose \methodname (Reflective Fault Localization Engine for Agent Correction). While prior work has used targeted intervention to validate debugging hypotheses~\cite{ma2025dover}, \methodname is the first to feed verified outcome flips back into step-level re-localization, closing the loop from correction to attribution. It localizes the earliest decisive error step in a completed, non-crashing trace through three stages: (1) \emph{diagnose} a candidate error step and produce a structured repair plan; (2) \emph{test} the hypothesis by injecting a targeted correction at that step and replaying the agent from the localized point, preserving the original prefix; and (3) \emph{verify and re-localize}, if the replayed trace produces the correct answer, the contrastive evidence between the original and corrected continuations sharpens the final attribution. The output is an \emph{attribution record}: the localized step, the intervention applied, and the verified outcome change.

Our primary evaluation regime assumes access to expected answers, modeling \emph{development-time debugging}: the natural setting for CI/CD pipelines, unit-tested workflows, and any task with deterministic or strongly checkable outputs where developers need to understand \emph{why} an agent failed, not just \emph{whether} it failed.

Our contributions are as follows:
\begin{enumerate}
    \item We identify four principal requirements for faithful error attribution in agent traces: execution grounding, prefix-preserving replay, targeted intervention, and inference-time computation.

    \item We propose \methodname, a method that instantiates all four requirements in a unified pipeline by combining diagnosis-guided intervention with controlled replay, producing \emph{attribution records} backed by intervention evidence rather than free-form rationales alone.

    \item We evaluate \methodname across four benchmarks spanning table question answering, multi-hop reasoning, chain-of-thought verification, and software engineering. \methodname achieves the highest localization accuracy among same-auditor methods, significantly improves localization when correction is successful, while still providing actionable localization even when ground truth answers are unavailable, with code and data provided in the supplementary material.\footnote{The \texttt{err-loc} library (core pipeline, all baselines, evaluation metrics) and the WTQ agent traces with human step-level annotations are included as supplementary material.}

\end{enumerate}
\section{Related Work}
\label{sec:related}

\paragraph{Inference-time correction methods:} reflection~\cite{shinn2023reflexion}, self-consistency~\cite{wang2022selfconsistency}, self-debug~\cite{chen2023selfdebug}, and extended reasoning~\cite{openai2024learning_reason_with_llms,guo2025deepseekr1,yao2023react,zhong2024debuglikehuman} improve task accuracy but do not diagnose \emph{why} the original trace failed; their critiques attach to new attempts rather than tested attributions of the original trajectory~\cite{huang2023selfcorrect,tyen2024mistakefinding,turpin2023cotunfaithful}.


\paragraph{Localization without execution grounding (violates R1).}
TRAIL~\cite{trail2025tracelocalization} provides a benchmark and taxonomy for span-level trace diagnosis but evaluates LLM judges that inspect traces without any execution-based verification.
AgenTracer~\cite{agentracer2025localization} uses counterfactual replay to \emph{curate training data}, then distills the signal into a fine-tuned 8B model that at inference time outputs a learned prediction without performing any intervention (also violates R4).
TrajAD~\cite{liu2026trajad} similarly trains a supervised anomaly detector on synthetically perturbed trajectories, producing step-level classifications without causal testing (also violates R4).
ThinkPRM~\cite{thinkprm2025} and related process reward models~\cite{choudhury2025agentprm,xi2025agentprm} score each step via a trained verifier, but the score reflects training-distribution patterns rather than test-time intervention evidence (also violates R4).

\paragraph{Signal checking without causal testing (violates R1, R2).}
AgentRx~\cite{agentrx2026} synthesizes constraints from tool schemas and audits each step for violations, producing structured evidence logs, but never tests whether fixing a flagged step actually changes the outcome.

\paragraph{Correction without attribution coupling (violates R2, R3).}
Reflexion~\cite{shinn2023reflexion} generates verbal self-critiques and retries, but the new trajectory is independent of the original prefix, so success does not identify which original step was decisive.
ICS~\cite{ics2025backtrack} backtracks to a localized step and resamples from a bare prefix, an untargeted retry that conflates stochastic variation with causal signal (violates R3), and does not use correction outcomes to refine attribution.
DoVer~\cite{ma2025dover} creates a failure hypothesis via an LLM log summarizer, then uses targeted intervention to validate whether the hypothesized fix resolves the failure. However, the replay outcome (Validated/Refuted/Inconclusive) is only used to evaluate the correction, and the localization never gets refined with intervention feedback.
LDB~\cite{zhong2024debuglikehuman} debugs code by verifying runtime execution step-by-step with breakpoint-based inspection~\cite{weiser1981programslicing,zeller2002delta}, but does not produce intervention-supported attribution records.

\methodname is, to our knowledge, the first method that closes the attribution loop by feeding intervention outcomes back into step-level re-localization, thereby combining all four properties in a unified pipeline. The contrast is summarized in Table~\ref{tab:method_comparison}. We provide extended per-baseline positioning in Appendix~\ref{sec:extended_related_work}.

\begin{table}[t]
\centering
\scriptsize
\caption{Comparison by attribution requirements. R1: execution-grounded (not narrative-only). R2: prefix-preserving replay (not independent retry). R3: targeted intervention (not bare resampling). R4: inference-time computation (not pre-trained classifier). \halfmark~denotes partial satisfaction. Extended justification in Appendix~\ref{sec:extended_related_work}.}
\label{tab:method_comparison}
\begin{tabular}{@{}lcccc@{}}
\toprule
Method & R1 & R2 & R3 & R4 \\
\midrule
Reflection~\cite{shinn2023reflexion} & \ding{55} & \ding{55} & \ding{55} & \checkmark \\
ICS~\cite{ics2025backtrack} & \halfmark & \checkmark & \ding{55} & \checkmark \\
LLM-as-judge~\cite{trail2025tracelocalization} & \ding{55} & \ding{55} & \ding{55} & \checkmark \\
AgenTracer~\cite{agentracer2025localization} & \ding{55} & \ding{55} & \ding{55} & \ding{55} \\
TrajAD~\cite{liu2026trajad} & \ding{55} & \ding{55} & \ding{55} & \ding{55} \\
ThinkPRM~\cite{thinkprm2025} & \ding{55} & \ding{55} & \ding{55} & \ding{55} \\
AgentRx~\cite{agentrx2026} & \ding{55} & \ding{55} & \ding{55} & \checkmark \\
DoVer~\cite{ma2025dover} & \halfmark & \halfmark & \checkmark & \checkmark \\
\methodname & \checkmark & \checkmark & \checkmark & \checkmark \\
\bottomrule
\end{tabular}
\end{table}
\section{Method: \methodname}
\label{sec:method}

\begin{figure*}[t]
\centering
\includegraphics[width=\textwidth]{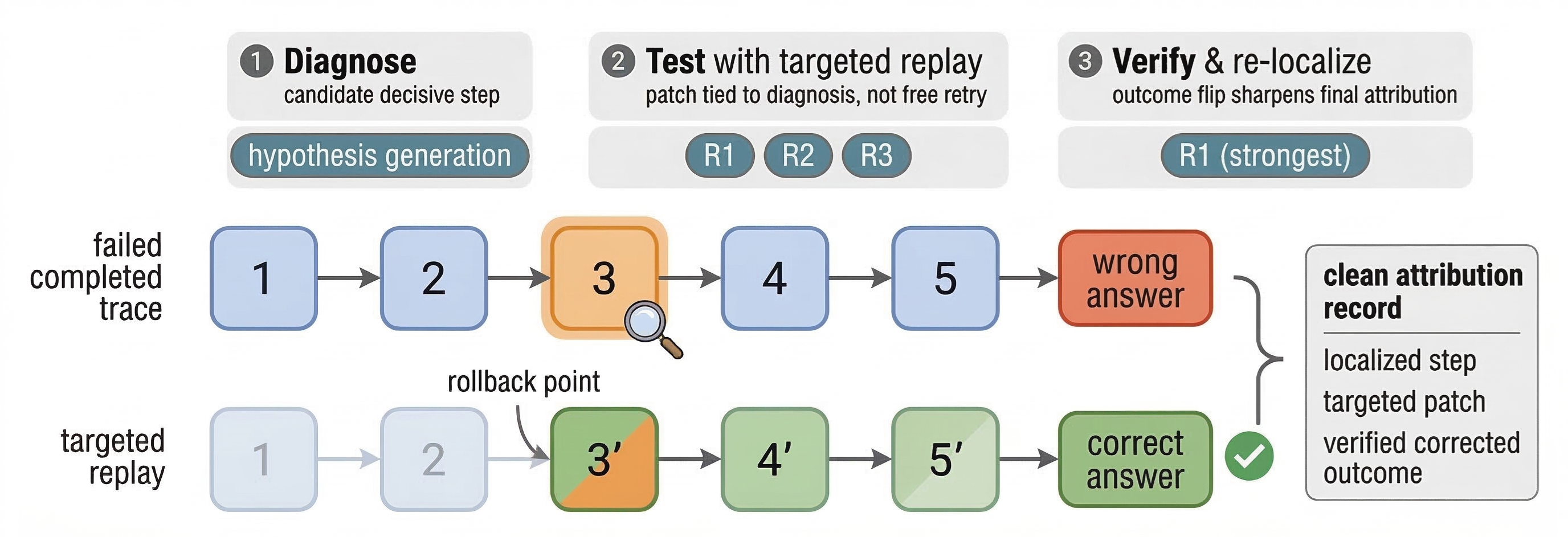}
\caption{\methodname in three stages, mapped to the four attribution requirements (R1--R4). Stage~1 produces a candidate error hypothesis. Stage~2 tests it through prefix-preserving targeted replay (R1, R2, R3). Stage~3 is the conceptual novelty: successful correction is reused as contrastive evidence to refine attribution rather than merely validating a fix. The verified rollback point becomes the attributed step index; the original and corrected continuations refine the error explanation at that point. R4 holds by construction.}
\label{fig:pipeline}
\end{figure*}

\methodname localizes the earliest decisive error step in a completed, non-crashing trace through three stages (Figure~\ref{fig:pipeline}): \emph{diagnose} a candidate step (\S\ref{sec:localization}), \emph{test} it through targeted replay (\S\ref{sec:intervene}), and \emph{verify and re-localize} using the outcome as contrastive evidence (\S\ref{sec:verify_certify}). All stages operate at inference time on the test trace (R4).

\subsection{Problem Setup and Attribution Record}
\label{sec:problem}

Let an agent trace be
\[
\tau = (u_1, \dots, u_T, y),
\]
where each step $u_t$ records the model-side state at step $t$ (reasoning or assistant message), the action taken if any, and the resulting observation if any. For tool-use agents, $u_t$ may be unpacked as $(s_t, a_t, o_t)$; for pure reasoning traces, it reduces to a reasoning step. We assume the run completed without runtime exceptions, yet the final answer $y$ is incorrect.

The localization target is the \emph{earliest decisive error step} $i^{*}$: the earliest step whose correction enables a verified-correct continuation under replay. We use this operational notion because traces may contain multiple correlated mistakes, while only some are decisive for the final failure.

An \emph{attribution record} is an operational object $(i^{*}, \Delta_{i^{*}})$ consisting of a targeted intervention at step $i^{*}$ and a controlled replay that (i)~preserves the prefix $\tau_{< i^{*}}$ and (ii)~flips the outcome from incorrect to correct.
When correction succeeds, this record provides intervention-supported evidence for the attribution; it establishes that the intervention at $i^{*}$ is \emph{sufficient} to change the outcome, though it does not claim causal minimality or uniqueness; other interventions may also suffice.
This operational definition directly implements the interventional criterion used in concurrent work for ground-truth annotation~\cite{agentracer2025localization}, but applied here as an inference-time test on each trace rather than a labeling protocol for training data.
When correction fails after all rounds, the record falls back to the best available localization from the iterative diagnosis process.

\subsection{Stage 1: Diagnosis}
\label{sec:localization}

Stage~1 produces the attribution hypothesis to be tested by the subsequent stages.
The localizer audits the full trace and predicts a candidate earliest decisive error step. Given the task input, the completed trace, the agent's final answer, and, in the oracle regime, the expected answer, an auditor LLM returns
\begin{equation}
\ell_0 = (\hat{\imath}_0,\; c), \label{eq:loc}
\end{equation}
where $\hat{\imath}_0$ is the provisional candidate step and $c \in [0,1]$ is a confidence score. Free-text reasoning and a suggested fix are also retained for downstream use, but the main object passed to later stages is the candidate step $\hat{\imath}_0$. All steps before $\hat{\imath}_0$ are treated as the preserved prefix. In the primary oracle regime, the expected answer may be provided to the localizer; in proxy settings it is withheld. Prompt formatting details are deferred to Appendix~A.

\subsection{Stage 2: Targeted Replay}
\label{sec:intervene}

Because replay is constrained by a diagnosis-specific repair plan rather than unconstrained retry (R3), validated empirically in \S\ref{sec:faithfulness}, and preserves the original prefix while producing an execution outcome, it simultaneously satisfies R1 (execution grounding), R2 (prefix-preserving replay), and R3.

Given a localized error step, Stage~2 constructs a targeted intervention and replays the agent from a nearby rollback point. The goal is not to obtain any successful retry, but to test whether changing the hypothesized step changes the downstream outcome. 

\paragraph{Error classification.}
\label{sec:classification}
The classifier maps the localized step to a coarse error family and produces a short correction sketch. This label is not itself the contribution; it serves only to make the subsequent repair plan more specific. The full taxonomy is deferred to Appendix~A.

\paragraph{Repair planning.}
\label{sec:repair_planning}
From the localized step and its coarse error family, the Diagnostician produces a structured repair plan
\begin{equation}
\pi = \big(\text{root\_cause},\; \text{instr},\; \text{forbidden},\; \text{tool}_{\text{exp}}\big), \label{eq:plan}
\end{equation}
containing a brief diagnosis, a correction instruction, actions to avoid, and an expected next tool when applicable. The purpose of $\pi$ is to constrain replay toward a targeted test of the hypothesized error rather than an unconstrained retry. Prompt templates and ablation-specific null plans are deferred to Appendix~A.

\paragraph{Rollback selection.}
\label{sec:rollback}
Because localization may be off by one or two steps, \methodname\ replays from a small neighborhood around $\hat{\imath}_0$:
\[
\{\hat{\imath}_0, \hat{\imath}_0{-}1, \dots, \max(\hat{\imath}_0{-}K{+}1, 1)\}.
\]
This lets the method recover from slight localization error while preserving as much of the original prefix as possible.

\paragraph{Repair loop.}
\label{sec:repair}
Algorithm~\ref{alg:repair} summarizes the replay procedure. In each diagnosis round, \methodname\ builds a repair plan, selects rollback points near the localized step, and reruns the agent from each prefix with injected guidance. The first regenerated step is optionally gated for faithfulness to the plan (\S\ref{sec:gate}), and a verifier determines whether the replayed trace corrects the outcome. On success, \methodname\ uses the corrected continuation to refine the attribution (\S\ref{sec:postfix}); on failure, it relocalizes on the best candidate trace and continues to the next round.
A successful correction is informative only when it is tied to a targeted intervention near the localized step; otherwise it is treated as an uninformative retry rather than evidence of attribution.

\begin{algorithm}[t]
\caption{\methodname Repair Loop}
\label{alg:repair}
\small
\begin{algorithmic}[1]
\Require Trace $\tau$, localization $\ell$, classification $\kappa$
\Ensure RepairResult (success/fail, corrected trace, final attribution)
\For{$d = 1, \dots, D$} \Comment{diagnosis rounds}
    \State $\pi_d \gets \textsc{Diagnose}(\tau, \ell, \kappa)$
    \State $P \gets \textsc{RollbackPoints}(\ell.\text{prefix\_len}, K)$
    \For{$p \in P$} \Comment{rollback points}
        \State $m \gets \textsc{BuildInjection}(\pi_d, p)$
        \State $\tau' \gets \textsc{AgentRerun}(\tau_{\le p}, m)$
        \State $v \gets \textsc{Gate}(\pi_d, \tau'_{p+1})$ \Comment{\S\ref{sec:gate}}
        \If{hard gate \textbf{and} $\neg v.\text{faithful}$}
            \State retry up to $R$ times with feedback
        \EndIf
        \If{$\textsc{Verify}(\tau') = \texttt{correct}$}
            \State $\ell' \gets \textsc{PostCorrExplain}(\tau, \tau', \ell)$
            \State \Return \texttt{success}, $\tau'$, $\ell'$
        \EndIf
    \EndFor
    \State $\tau_{\text{best}} \gets \textsc{SelectBest}(\text{candidates})$ \Comment{best-fix score (proxy) / failure diagnosis (oracle)}
    \State $\ell \gets \textsc{Localize}(\tau_{\text{best}})$ \Comment{re-localize}
    \State $\kappa \gets \textsc{Classify}(\tau_{\text{best}}, \ell)$
    \State $\tau \gets \tau_{\text{best}}$
\EndFor
\State \Return \texttt{fail}, $\tau$, $\ell$
\end{algorithmic}
\end{algorithm}

\subsubsection{Faithfulness Gate}
\label{sec:gate}

The faithfulness gate checks whether the first regenerated step follows the repair plan rather than drifting to an unrelated recovery path~\cite{jacovi2020faithfulness,doshivelez2017interpretable}. We implement this as a lightweight rule-based filter plus an LLM judgment when needed. In soft mode the verdict is logged; in hard mode unfaithful first steps are rejected and replay is retried with feedback.
The gate operationalizes the distinction between targeted intervention and uninformative retry~(R3): if the agent ignores the repair plan and succeeds via an unrelated recovery path, that success does not constitute evidence about the hypothesized error step.
The exact verdict schema and retry policy are deferred to Appendix~A.

\subsection{Stage 3: Verified Rollback Attribution and Explanation Refinement}
\label{sec:verify_certify}

Stage~3 performs two distinct functions. First, \emph{verified rollback-point attribution}: the rollback point that yields the outcome flip is adopted as the attributed step index, since it is the point for which there is direct intervention evidence. Second, \emph{contrastive explanation refinement}: the original and corrected continuations at that point are compared to sharpen the error explanation. Methods that use intervention only to validate a fix---classifying hypotheses as supported or refuted without revising the localized step (e.g., DoVer; \citealp{ma2025dover})---stop after the first function; \methodname additionally re-diagnoses from failed candidate traces and uses a faithfulness gate (§\ref{sec:gate}) so that outcome flips count as attribution evidence only when tied to the diagnosis-specific repair plan.
In the primary oracle regime, a replay that changes the final answer from incorrect to correct completes an attribution record for the intervened step.

\paragraph{Verification signal.}
\label{sec:verification}
When the expected answer $y^{*}$ is available, verification reduces to task-specific answer matching between the replayed trace and $y^{*}$. This is the clean supervision signal used in the main experiments.

\paragraph{Contrastive explanation refinement.}
\label{sec:postfix}
Once the verified rollback point sets the step index, a post-correction step compares the original and corrected continuations at that point to refine the error explanation: the error type, root cause description, and free-text reasoning are updated based on what changed between the two trajectories. The step index itself is determined entirely by which rollback point yielded the outcome flip---not by the contrastive comparison. The output is a \emph{sufficient intervention point} that may be at or near, but not necessarily identical to, the earliest causal origin; single-step attribution may be ill-posed in traces with distributed or multi-trial failures, which we scope as a limitation (§\ref{sec:limitations}).

\paragraph{Final attribution record.}
The output of Stage~3 is an attribution record $(\hat{\imath}_{\text{final}},\; \Delta_{\hat{\imath}_{\text{final}}})$, where $\hat{\imath}_{\text{final}}$ is the re-localized step index after contrastive explanation and $\Delta_{\hat{\imath}_{\text{final}}}$ is the intervention that flipped the outcome. When correction fails after all rounds, the record falls back to the best available localization from the last round.

\paragraph{Proxy verification (secondary regime).}
\label{sec:plan_anchor_verifier}
When $y^{*}$ is unavailable, we replace the oracle with a \emph{plan-anchored verifier} that estimates correctness without ground truth. The verifier synthesizes task anchors from the question, audits each macro-step for deviations, and outputs a calibrated error signal. This regime supports deployment-time localization but is weaker than oracle-verified attribution; the full architecture and detection results are in Appendix~\ref{app:plan_anchor_verifier}.

\section{Experiment}
\label{sec:experiment}

Our experiments are organized around the four attribution requirements. RQ1 (\S\ref{sec:localization_results}) tests whether methods lacking execution grounding~(R1), including the strongest available LLM judge, can match intervention-based attribution. RQ2 (\S\ref{sec:correction_coupling}) tests whether \methodname's correction is coupled to understanding, the operational signature of R1 and R2 working together. RQ3 (\S\ref{sec:faithfulness}) directly tests R3 by comparing targeted intervention against bare resampling and semantic controls. RQ4 (\S\ref{sec:proxy_results}) extends the framework to settings without oracle verification. The ablation (\S\ref{sec:ablation}) isolates the contribution of each \methodname component to the requirements it serves.

\subsection{Experimental Setup}

\paragraph{Datasets.}
We evaluate on four benchmarks of failing agent traces:
\textbf{WTQ} (table-QA): we run a tabular-QA agent on a subset of WikiTableQuestions~\cite{pasupat2015wtq} and collect step-level human annotations on the failing traces;
\textbf{GAIA} (multi-hop reasoning) and \textbf{SWE-bench} (software engineering): we use the agent traces and human step-level annotations released by the TRAIL benchmark~\cite{trail2025tracelocalization}, which were generated over GAIA~\cite{mialon2023gaia} and SWE-bench~\cite{jimenez2023swebench} tasks respectively;
\textbf{BBM}~\cite{tyen2024mistakefinding} (chain-of-thought verification): serves as a boundary stress-test, as its traces lack tool-call structure and the intervention signal is weakest.
Table~\ref{tab:data_summary} summarizes trace counts.

\begin{table}[h]
\centering
\small
\caption{Dataset summary. Human-labeled traces have step-level error annotations.}
\label{tab:data_summary}
\begin{tabular}{@{}lrr@{}}
\toprule
Benchmark & Traces & Human-labeled \\
\midrule
WTQ & 137 & 119 \\
GAIA & 117 & 83 \\
BBM & 150 & 150 \\
SWE-bench & 31 & 30 \\
\bottomrule
\end{tabular}
\end{table}

Each benchmark uses a task-appropriate agent implementation.
Agent configurations, tool-set details, annotation provenance, and inter-annotator agreement are in Appendix~\ref{app:agent_configs} and~\ref{app:iaa}.

\paragraph{Model, verification, and scoring.}
All experiments use \texttt{gpt-5.2-2025-12-11} as both auditor and agent (temperature 0.0) unless otherwise noted.
Correctness is verified deterministically: string matching for WTQ and GAIA, task-specific extraction for BBM, and the benchmark test suite for SWE-bench (Appendix~\ref{app:verifier}).
In the primary oracle regime, the expected answer is available to all pipeline components including the localizer and diagnostician, modeling development-time debugging where the correct output is known.
The proxy regime (\S\ref{sec:proxy_results}) removes this assumption entirely.
Error step predictions are evaluated via set-membership over all annotators' unique error steps, with standard errors in parentheses.

\paragraph{Baselines.}
We compare against eight baselines spanning four paradigms:
(i)~\emph{prompt-based}: AAO, SBS, and BS, each using the same auditor LLM;
(ii)~\emph{correction-based}: ICS~\cite{ics2025backtrack} and Reflexion~\cite{shinn2023reflexion} (with and without ground truth);
(iii)~\emph{scoring/constraint}: AgentRx~\cite{agentrx2026} and ThinkPRM~\cite{thinkprm2025};
(iv)~\emph{correction-validation}: DoVer~\cite{ma2025dover}.
All baselines are described in \S\ref{sec:related}; algorithmic details are in Appendix~\ref{app:baselines}.

\subsection{Metrics}

\textbf{Localization accuracy (primary).}
\emph{Exact Match (EM)}: predicted step $\in$ human-annotated error step set.
\emph{Off-by-1}: predicted step within distance 1 of any annotated step.
\emph{Mean Distance}: average absolute distance to the nearest annotated step.

\textbf{Explanation quality.}
LLM-judge semantic similarity (0--1) between the auditor's free-text explanation and human error descriptions, scored only when the predicted step matches a human-annotated step (step-conditioned).
\emph{Expl.\ High}: fraction with similarity $\geq 0.7$.
Evaluated on WTQ, GAIA, and SWE-bench; BBM provides only \texttt{mistake\_index} without descriptions.
We also report a coverage-aware variant $\text{Sim}_{\text{cov}} = \text{Sim} \times (n_{\text{matched}} / n_{\text{total}})$, which assigns similarity~0 to unmatched predictions and accounts for differences in localization accuracy across methods (Appendix~\ref{app:explanations}).

\textbf{Correction rate.}
Fraction of traces where the repair loop produces the correct final answer.

\begin{table*}[!ht]
\centering
\small
\caption{Localization accuracy across benchmarks. EM is the primary metric; Off-1 is a softer localization measure. Standard errors in parentheses. AAO~(Opus~4.6) uses Claude Opus~4.6 as the auditor; all other rows use gpt-5.2. Proxy results are discussed in \S\ref{sec:proxy_results}.}
\label{tab:localization}
\begin{tabular}{@{}l cc cc cc cc@{}}
\toprule
& \multicolumn{2}{c}{WTQ} & \multicolumn{2}{c}{GAIA} & \multicolumn{2}{c}{BBM} & \multicolumn{2}{c}{SWE-bench} \\
\cmidrule(lr){2-3} \cmidrule(lr){4-5} \cmidrule(lr){6-7} \cmidrule(lr){8-9}
Method & EM & Off-1 & EM & Off-1 & EM & Off-1 & EM & Off-1 \\
\midrule
AAO & 55.5\tss{4.6} & 60.5\tss{4.5} & 15.7\tss{4.0} & 62.7\tss{5.3} & 30.3\tss{3.9} & 47.2\tss{4.2} & 33.3\tss{8.6} & 83.3\tss{6.8} \\
AAO (Opus 4.6) & 70.8\tss{4.2} & 76.6\tss{3.9} & 15.4\tss{4.0} & 49.6\tss{5.5} & 60.1\tss{4.1} & 70.6\tss{3.8} & 19.4\tss{7.6} & 77.4\tss{8.0} \\
SBS & 50.4\tss{4.6} & 58.0\tss{4.5} & 1.2\tss{1.2} & 34.9\tss{5.2} & 0.7\tss{0.7} & 26.1\tss{3.7} & 26.7\tss{8.1} & 80.0\tss{7.3} \\
BS & 53.8\tss{4.6} & 60.5\tss{4.5} & 13.3\tss{3.7} & 48.2\tss{5.5} & 16.2\tss{3.1} & 29.6\tss{3.8} & 33.3\tss{8.6} & 76.7\tss{7.7} \\
Reflexion+GT & 52.6\tss{4.6} & 67.9\tss{4.3} & 16.2\tss{4.0} & 47.9\tss{5.5} & 34.0\tss{4.0} & 52.7\tss{4.2} & 45.2\tss{9.1} & 87.1\tss{6.1} \\
Reflexion--noGT & 53.3\tss{4.6} & 62.8\tss{4.4} & 17.9\tss{4.2} & 45.3\tss{5.5} & 14.0\tss{2.9} & 33.3\tss{4.0} & 41.9\tss{9.0} & 83.9\tss{6.7} \\
ThinkPRM & 27.7\tss{4.1} & 59.9\tss{4.5} & 14.5\tss{3.9} & 35.0\tss{5.2} & 14.7\tss{3.0} & 34.7\tss{4.0} & 38.7\tss{8.9} & 80.6\tss{7.2} \\
AgentRx & 51.1\tss{4.6} & 68.6\tss{4.3} & 6.8\tss{2.8} & 31.6\tss{5.1} & 3.3\tss{1.5} & 29.3\tss{3.8} & 41.9\tss{9.0} & 83.9\tss{6.7} \\
ICS & 48.7\tss{4.6} & 54.6\tss{4.6} & 10.8\tss{3.4} & 43.4\tss{5.4} & 19.7\tss{3.3} & 31.7\tss{3.9} & 18.2\tss{7.0} & 90.9\tss{5.3} \\
DoVer+GT & 35.0\tss{4.4} & 52.6\tss{4.6} & 16.2\tss{4.0} & 32.5\tss{5.1} & 2.0\tss{1.2} & 25.3\tss{3.6} & 12.9\tss{6.1} & 80.6\tss{7.2} \\
\midrule
\textbf{\methodname} & \textbf{76.3}\tss{3.9} & \textbf{82.2}\tss{3.5} & \textbf{39.0}\tss{4.3} & \textbf{66.3}\tss{5.2} & \textbf{34.5}\tss{4.0} & \textbf{51.8}\tss{4.2} & \textbf{70.0}\tss{8.4} & \textbf{93.3}\tss{4.6} \\
\methodname (proxy) & 62.2\tss{4.5} & 75.6\tss{4.0} & 33.3\tss{5.4} & 68.2\tss{5.7} & 31.0\tss{3.9} & 45.1\tss{4.2} & 66.1\tss{9.4} & 80.0\tss{7.6} \\
\bottomrule
\end{tabular}
\end{table*}

\subsection{RQ1: Can Correction-Only or Judge-Only Methods Faithfully Localize?}
\label{sec:localization_results}

Table~\ref{tab:localization} presents error localization accuracy across all four benchmarks.

\methodname achieves the highest exact match among same-auditor methods on all four benchmarks, with the largest gains on WTQ and SWE-bench where post-correction re-localization provides strong contrastive evidence.
Replacing the auditor with Claude Opus~4.6 does not close the gap on tool-use benchmarks (70.8\% WTQ, 19.4\% SWE-bench), though Opus surpasses \methodname on BBM (60.1\% vs.\ 34.5\%), consistent with its boundary stress-test role.
Appendix~\ref{app:verified_fallback} stratifies localization by correction outcome; Appendix~\ref{app:cost} confirms gains are not attributable to higher compute.

\subsection{RQ2: Is Correction Coupled to Localization?}
\label{sec:correction_coupling}

The question is not whether \methodname can repair a trace, but whether successful repair is evidence of understanding.
Table~\ref{tab:correction} tests this by comparing correction rates and stratifying explanation quality by correction outcome.

\begin{table}[t]
\centering
\footnotesize
\caption{Correction--localization coupling. Sim-c / Sim-f denote explanation similarity for corrected / failed traces.}
\label{tab:correction}
\setlength{\tabcolsep}{3pt}
\begin{tabular}{@{}llcccc@{}}
\toprule
Bench. & Method & Corr.\% & Sim-c & Sim-f & $\Delta$ \\
\midrule
\multirow{3}{*}{WTQ}
  & ICS         & 23.7\tss{3.6} & 0.22\tss{.14} & 0.19\tss{.14} & +0.03\tss{.03} \\
  & Refl.       & 27.4\tss{3.9} & 0.25\tss{.15} & 0.20\tss{.14} & +0.05\tss{.04} \\
  & \methodname & 92.6\tss{2.2} & 0.61\tss{.16} & 0.32\tss{.18} & +0.29\tss{.06} \\
\midrule
\multirow{3}{*}{GAIA}
  & ICS         & 12.0\tss{3.0} & 0.18\tss{.12} & 0.15\tss{.11} & +0.03\tss{.03} \\
  & Refl.       & 15.6\tss{3.4} & 0.21\tss{.13} & 0.16\tss{.12} & +0.05\tss{.03} \\
  & \methodname & 54.2\tss{4.6} & 0.56\tss{.18} & 0.29\tss{.17} & +0.27\tss{.03} \\
\midrule
\multirow{3}{*}{SWE}
  & ICS         & 85.2\tss{6.4} & 0.13\tss{.10} & 0.11\tss{.09} & +0.02\tss{.04} \\
  & Refl.       & 86.8\tss{6.0} & 0.16\tss{.11} & 0.12\tss{.10} & +0.04\tss{.04} \\
  & \methodname & 89.0\tss{5.6} & 0.48\tss{.19} & 0.23\tss{.15} & +0.25\tss{.09} \\
\bottomrule
\end{tabular}
\end{table}
The signature result is the coupling gap~$\Delta$: \methodname shows a large explanation quality difference between corrected and failed traces ($\Delta = +0.25$ to $+0.29$), while both ICS ($\Delta \leq +0.03$) and Reflexion ($\Delta \leq +0.05$) show only small gaps despite non-trivial correction rates.
For \methodname, successful correction is accompanied by substantially better explanations (Sim$_{\text{c}}$ = 0.48--0.61 vs.\ Sim$_{\text{f}}$ = 0.23--0.32), validating the post-correction re-localization design (\S\ref{sec:postfix}).
This coupling is a direct consequence of prefix-preserving replay~(R2): the original and corrected traces share a prefix but diverge at the hypothesized step, providing structurally constrained contrastive evidence that independent retries cannot produce.

\subsection{RQ3: Is the Intervention Signal Faithful?}
\label{sec:faithfulness}

\begin{table}[t]
\centering
\small
\caption{Selected faithfulness results on 119 WTQ examples. Correct-hint variants outperform semantic controls and the no-hint baseline in both correction accuracy and plan adherence. We report the most diagnostic conditions here; the full condition set appears in Appendix~\ref{app:faithfulness}.}
\label{tab:faithfulness}
\begin{tabular}{@{}lcc@{}}
\toprule
Condition & Acc.\ (\%) & Adh.\ (\%) \\
\midrule
Paraphrased hint & \textbf{80.7}\tss{3.2} & 60.0\tss{4.0} \\
Correct hint & 78.7\tss{3.3} & 60.7\tss{4.0} \\
Position-matched hint & 77.0\tss{3.5} & 61.5\tss{4.0} \\
\midrule
Placebo hint & 46.3\tss{4.1} & 21.5\tss{3.4} \\
Contradictory hint & 39.2\tss{4.0} & 18.2\tss{3.2} \\
No hint & 35.6\tss{3.9} & 20.1\tss{3.3} \\
\midrule
Wrong-step intervention & 46.2\tss{9.8} & 60.3\tss{9.6} \\
\bottomrule
\end{tabular}
\end{table}

Table~\ref{tab:faithfulness} directly tests R3 (targeted intervention) by comparing correction accuracy and plan adherence under targeted hints versus bare resampling and semantic controls (119 WTQ examples, 3 samples per condition, temperature 0.2; full conditions in Appendix~\ref{app:faithfulness}).
Injecting the correct repair hint yields $+42.0$pp correction accuracy over no-hint and $+32.0$pp over placebo, confirming that the agent responds to the \emph{semantic content} of the intervention, not merely extra context tokens.
The effect is stable across paraphrase and position-matched variants (all within 6pp).
Intervening at a deliberately wrong step (mean distance 2.6) drops accuracy to 46.2\% while adherence stays high (60.3\%), confirming that the \emph{localization target}, not the act of intervention itself, drives correction.
Appendix~\ref{app:faithfulness} further reports immediate probability shifts and a $+13$pp correction gain when the agent adheres to high-quality plans.

\subsection{RQ4: What Survives Without Oracle Verification?}
\label{sec:proxy_results}
\methodname (proxy) replaces the oracle verifier with plan-anchored stepwise auditing (\S\ref{sec:plan_anchor_verifier}), synthesizing task anchors from the question alone.
Table~\ref{tab:localization} (last row) shows the proxy retains the majority of oracle accuracy (62.2\% WTQ, 66.1\% SWE-bench, 33.3\% GAIA).
The oracle-to-proxy gap reflects verifier quality (Table~\ref{tab:verifier_comparison}): AUROC reaches 0.97 on SWE-bench but only 0.56 on BBM, where chain-of-thought traces lack the tool-call structure that anchor-based auditing exploits.
Even with imperfect verification, proxy \methodname outperforms most oracle-aided baselines on WTQ and SWE-bench.

\begin{table}[t]
\centering
\small
\caption{Plan-anchored verifier detection performance. AUROC measures discrimination between correct and incorrect traces; 0.5 is chance.}
\label{tab:verifier_comparison}
\begin{tabular}{@{}lrrr@{}}
\toprule
Benchmark & AUROC & Best-F1 & Err-Prec \\
\midrule
WTQ        & 0.680 & 0.769 & 0.625 \\
BBM        & 0.561 & 0.702 & 0.568 \\
GAIA       & 0.842 & 0.888 & 0.934 \\
SWE-bench  & 0.968 & 0.933 & 0.966 \\
\bottomrule
\end{tabular}
\end{table}

\subsection{Ablation Study}
\label{sec:ablation}

We measure the contribution of each \methodname component by running seven ablation configurations on WikiTableQuestions, toggling Diagnostician, Hard Gate, and Multi-rollback (Appendix~\ref{app:ics}).
All configs share the same localizer (all-at-once with expected answer) and model.

\begin{table}[!h]
\centering
\footnotesize
\caption{Ablation on WTQ. Diag = Diagnostician, Gate = Hard Gate, RB = rollback points. Configs 1--7 report EM before post-correction re-localization; the full \methodname row includes it.}
\label{tab:ablation}
\begin{tabular}{@{}clcccrr@{}}
\toprule
\# & Config & Diag & Gate & RB & EM (\%) & Corr.\ (\%) \\
\midrule
1 & bare & $\times$ & $\times$ & 1 & 62.8\tss{4.4} & 54.7\tss{4.3} \\
2 & + Diag & $\checkmark$ & $\times$ & 1 & 63.5\tss{4.4} & 83.7\tss{3.2} \\
3 & + Gate & $\times$ & $\checkmark$ & 1 & 64.2\tss{4.4} & 55.6\tss{4.2} \\
4 & + Multi & $\times$ & $\times$ & 3 & 66.4\tss{4.3} & 62.2\tss{4.1} \\
5 & + D\,+\,G & $\checkmark$ & $\checkmark$ & 1 & 62.8\tss{4.4} & 82.7\tss{3.2} \\
6 & + D\,+\,M & $\checkmark$ & $\times$ & 3 & 64.2\tss{4.4} & 89.5\tss{2.6} \\
7 & -- re-loc & $\checkmark$ & $\checkmark$ & 3 & 65.7\tss{4.4} & 92.6\tss{2.2} \\
\midrule
& \textbf{\methodname} & $\checkmark$ & $\checkmark$ & 3 & \textbf{76.3}\tss{3.9} & 92.6\tss{2.2} \\
\bottomrule
\end{tabular}
\end{table}

The Diagnostician is the most impactful component for correction ($\overline{\Delta}$Corr: $+27.8$pp); multi-rollback drives localization ($\overline{\Delta}$EM: $+2.2$pp); the hard gate safeguards intervention specificity~(R3) with minimal EM cost.
Post-correction re-localization provides a further $+10.6$pp boost (65.7\% $\to$ 76.3\% EM), confirming that the attribution feedback loop, not correction alone, distinguishes \methodname from correction-only methods.

\begin{table}[t]
\centering
\small
\caption{Localization stratified by correction outcome. Cov.\% = correction rate. Verified = intervention-supported attribution; Fallback = Stage-1 hypothesis only.}
\label{tab:verified_fallback}
\begin{tabular}{@{}lrcccc@{}}
\toprule
Benchmark & Cov.\% & \multicolumn{2}{c}{Verified} & \multicolumn{2}{c}{Fallback} \\
\cmidrule(lr){3-4} \cmidrule(lr){5-6}
& & EM & Off-1 & EM & Off-1 \\
\midrule
WTQ       & 92.6 & 74.1 & 80.6 & 100.0 & 100.0 \\
GAIA      & 54.2 & 47.1 & 72.5 &  25.8 &  74.2 \\
BBM       & 95.0 & 35.8 & 53.7 &  14.3 &  42.9 \\
SWE-bench & 89.0 & 77.7 & 75.0 &   0.0 & 100.0 \\
\bottomrule
\end{tabular}
\end{table}
Table~\ref{tab:verified_fallback} asks whether Stage~3 verified attribution adds localization value beyond Stage-1 alone. On GAIA ($+21.3$pp) and BBM ($+21.5$pp), verified cases substantially outperform fallback, confirming that intervention evidence yields stronger attribution than pattern-matching alone (WTQ fallback: $N{=}10$, too small to interpret).

\section{Conclusion}

\methodname closes the loop from correction back to attribution by diagnosing a candidate error step, testing it through targeted replay, and using the verified outcome flip as contrastive evidence to refine the final localization.
Across four benchmarks spanning table-QA, multi-hop reasoning, chain-of-thought, and software engineering, \methodname achieves the highest same-auditor localization accuracy and exhibits a strong correction--localization coupling absent in all baselines. The method's gains are largest on structured tool-use traces and more modest on pure chain-of-thought reasoning where the intervention signal is weaker.
Future work should target stronger no-oracle verification and replay mechanisms for unstructured traces.

\section*{Acknowledgements}
This work was supported in part by a gift from Cisco, the National Science Foundation (NSF CNS-2247795), and the Office of Naval Research (ONR N00014-22-1-2680).

\bibliography{agent}
\bibliographystyle{icml2026}

\appendix

\section{Prompt Templates}
\label{app:prompts}

This section provides the verbatim prompt templates used in all pipeline stages.
All prompts require the LLM to return structured JSON; free-text fields (\texttt{reasoning}, \texttt{explanation}) are capped at the model's natural response length.
The prompts shown below use WTQ-specific formatting for concreteness. Benchmark-specific prompt variants for GAIA, BBM, and SWE-bench are included in the submitted \texttt{err-loc} library; the pipeline structure and JSON schema are shared across all benchmarks, with differences limited to domain-specific field names (e.g., ``table context'' for WTQ vs.\ ``task description'' for GAIA) and tool-name references.

\subsection{Error Localization Prompt}
\label{app:prompt_loc}

Used by \texttt{ErrorLocalizer} (\S\ref{sec:localization}).
The numbered steps are formatted with tool name, arguments, reasoning, and truncated result (max 500 chars per field).

\begin{small}
\begin{verbatim}
You are an expert auditor analyzing
a data-analysis agent trace.

The trace is known to be incorrect.

Question: {question}

Table context:
{table_context}

Expected answer: {expected_answer}
Agent final answer: {agent_answer}

Numbered steps:
{numbered_steps}

Identify the FIRST step where the error
was introduced.
Return JSON only:
{
  "error_step": <integer>,
  "confidence": <0.0-1.0>,
  "reasoning": "<short explanation>",
  "what_should_have_been_done":
    "<short correction>"
}
\end{verbatim}
\end{small}

When \texttt{include\_expected\_answer=False}, the \texttt{expected\_answer} field is replaced with the literal string \texttt{[WITHHELD]}.

\subsection{Error Classification Prompt}
\label{app:prompt_cls}

Used by \texttt{ErrorClassifier} (\S\ref{sec:classification}).
The error step is presented with its tool name, arguments, reasoning, and result.

\begin{small}
\begin{verbatim}
You are classifying the first error step
in a tool-calling trace.

Question: {question}
Expected answer: {expected_answer}
Agent answer: {agent_answer}

Error step {step_index}:
Tool: {tool_name}
Reasoning: {ai_reasoning}
Arguments: {tool_args}
Result: {tool_result}

Classify into one of:
- formatting_error
- instruction_noncompliance
- context_handling_failure
- resource_abuse
- poor_information_retrieval
- incorrect_problem_identification
- hallucination
- tool_error
- task_orchestration
- goal_deviation
- incorrect_memory_usage

Return JSON only:
{
  "error_type": "<category>",
  "confidence": <0.0-1.0>,
  "explanation": "<short explanation>",
  "suggested_correction": "<short fix>"
}
\end{verbatim}
\end{small}

\subsection{Diagnosis Prompt}
\label{app:prompt_diag}

Used by \texttt{Diagnostician} (\S\ref{sec:repair_planning}).
The prompt presents the correct prefix (all steps before the error), the erroneous step with full context, and the error cascade (all steps from the error onward).

\begin{small}
\begin{verbatim}
You are a diagnostic expert analyzing why
a data-analysis agent failed.

**Question:** {question}
**Available Data:** {table_context}
**Expected Correct Answer:**
  {expected_answer}
---
**Correct Steps (before the error):**
{correct_prefix}

**The Erroneous Step:**
{error_step}

**What Happened After (cascade):**
{error_cascade}

**Error Classification:** {error_type}
**Classification Explanation:**
  {classification_explanation}
**Previously Suggested Correction:**
  {suggested_correction}
---
Produce a repair plan:
1. ROOT CAUSE: Why the agent chose wrongly
2. What EXACTLY to do instead
3. What the agent must NEVER do
4. expected_next_tool: first tool token
   (e.g., f_sort_by), or null

Respond in JSON:
{
  "root_cause": "<explanation>",
  "correction_instruction":
    "<specific instruction>",
  "forbidden_actions": ["<action>", ...],
  "expected_next_tool": "<tool or null>",
  "confidence": <0.0-1.0>
}
\end{verbatim}
\end{small}

\subsection{Faithfulness Gate Prompt}
\label{app:prompt_gate}

Used by \texttt{FaithfulnessGate} (\S\ref{sec:gate}) after rule-based checks pass.

\begin{small}
\begin{verbatim}
You are a gatekeeper verifying that an
agent's proposed action follows a repair
plan.

**Repair Plan:**
- Correction Instruction:
    {correction_instruction}
- Forbidden Actions: {forbidden_actions}
- Expected Next Tool:
    {expected_next_tool}

**Proposed Action:**
- Tool: {proposed_tool}
- Arguments: {proposed_args}
- Agent's Reasoning: {proposed_reasoning}
---
Check:
1. Does it follow the correction?
2. Does it violate any forbidden action?
3. Is it relevant, or has the agent
   drifted?

Respond in JSON:
{
  "is_faithful": <true or false>,
  "violation_reason": "<or null>",
  "violation_type": "<one of:
    repeats_original_error,
    ignores_instruction,
    uses_forbidden_action,
    unrelated_drift, or null>"
}
\end{verbatim}
\end{small}

\subsection{Repair System Injection}
\label{app:prompt_inject}

Injected as a system message into the agent's context at the rollback point.
This is not an LLM prompt per se but a structured message appended to the conversation prefix.

\begin{small}
\begin{verbatim}
IMPORTANT - CORRECTION CONTEXT:
Your previous attempt to answer this
question failed at Step {error_step_index}.

Diagnosis: {root_cause}

You MUST follow this correction:
{correction_instruction}

You MUST NOT:
{forbidden_actions_formatted}

Resume from the last correct state and
apply this correction.
\end{verbatim}
\end{small}

When hard gate retries are triggered, feedback from the previous rejection is appended:

\begin{small}
\begin{verbatim}
GATE RETRY FEEDBACK (retry N failed):
- Previous violation_type: {type}
- Previous violation_reason: {reason}
- Previous first regenerated tool: {tool}
- Previous first regenerated args: {args}
- Regenerate Step {index} again and do
  not repeat the same first action.
- First regenerated tool must satisfy
  expected_next_tool={expected_tool}.
\end{verbatim}
\end{small}

\subsection{Post-Correction Explanation Prompt}
\label{app:prompt_postfix}

Used by \texttt{post\_correction\_explain} (\S\ref{sec:postfix}) after a successful correction.

\begin{small}
\begin{verbatim}
You are an expert auditor explaining why
a specific step in a data-analysis agent
trace was wrong.

A correction loop rolled back to step
{error_step_index} and resampled. The
resampled continuation produced the
CORRECT answer.

**Question:** {question}
**Expected answer:** {expected_answer}

**Original error step ({error_step_index}):**
Tool: {original_tool}
Reasoning: {original_reasoning}
Arguments: {original_args}
Result: {original_result}

**Resampled continuation:**
{resampled_steps}

**Corrected answer:** {corrected_answer}

Explain WHY the original step was wrong,
using the correction as evidence.

Return JSON only:
{
  "reasoning": "<detailed explanation>",
  "error_type": "<category>",
  "explanation": "<short summary>"
}
\end{verbatim}
\end{small}

\paragraph{Step-index determination.}
The re-localized step index $\hat{\imath}_{\mathrm{final}}$ is not produced by this prompt.
When correction succeeds from rollback point~$p$, the re-localized step is set to the rollback point that yielded the successful correction (i.e., $\hat{\imath}_{\mathrm{final}} = p$).
The post-correction explanation prompt produces the refined error explanation and error type for that step; the step index is determined mechanically by the replay infrastructure, not by LLM judgment.

\section{Baseline Localizer Algorithms}
\label{app:baselines}

All three baselines share the same input format: a question, expected answer, and a list of formatted failure log entries (one per step).
Each entry includes the step index, tool name, arguments, reasoning, and result.

\subsection{All-at-Once (AAO)}

The simplest baseline: a single LLM call receives the full trace and must identify the earliest decisive error step.
One LLM call per trace.

The prompt asks the LLM to identify the ``earliest decisive error step,'' defined as the step that, if fixed, would allow the run to succeed.
The LLM returns the error step index, responsible agent, reason, error type, and confidence.

\subsection{Step-by-Step (SBS)}

The LLM evaluates each step sequentially.
For step $t$, the prompt presents all steps $1, \dots, t$ and asks whether step $t$ is the first decisive error.
The LLM returns \texttt{is\_error: true/false}.
The first step marked as an error is returned; if no step is flagged, the final step is used as fallback.
$T$ LLM calls per trace in the worst case.

\subsection{Binary Search (BS)}

The trace is recursively bisected.
At each iteration, the LLM sees the current range $[\text{low}, \text{high}]$ split into two halves and must choose which half contains the earliest decisive error.
After convergence to a single step, a follow-up prompt generates a reason and error type.
$\lceil \log_2 T \rceil + 1$ LLM calls per trace.

\section{ICS Backtracker Details}
\label{app:ics}

Our ICS implementation follows \citet{ics2025backtrack}.
The agent resamples from a bare prefix at the localized step without any error hints or structured repair guidance, and the localizer does not see the expected answer.
If the resample produces a correct answer, the trace is recorded; otherwise, the method re-localizes on the new trace and repeats for up to $D$ iterations.
This matches the untargeted resampling configuration described in \citet{ics2025backtrack}: no diagnosis, no repair plan, no faithfulness gate.
ICS uses a single fixed rollback point per iteration (no multi-rollback).
Full implementation details are available in the submitted \texttt{err-loc} library.

\section{Verification}
\label{app:verifier}

Four verifier implementations are used depending on the benchmark:

\paragraph{OracleVerifier.}
Used for WTQ and GAIA.
A deterministic string-extraction pipeline extracts the agent's final answer from its output, normalizes formatting (case, whitespace, number formatting), and compares against the expected answer via string matching.
If extraction fails (no parseable answer string), the trace is marked incorrect.
This approach is intentionally strict: it may under-count correct answers when the agent uses equivalent but differently formatted expressions, but ensures verification introduces no LLM judgment uncertainty.
Returns \texttt{is\_correct} and the extracted answer string.

\paragraph{BBMVerifier.}
Used for BBM.
Extracts the answer from the agent's final output using a task-specific extractor (handles multiple-choice and short-answer formats), then performs case-insensitive string comparison.
No LLM call.

\paragraph{SWEVerifier.}
Used for SWE-bench.
Executes the benchmark's test suite against the agent's generated patch.
Returns pass/fail based on whether all tests pass.
No LLM call is involved.

\paragraph{SelfVerifier.}
Available for use when ground-truth answers are unavailable.
An LLM receives the question, table context, agent reasoning summary, and agent answer, then judges whether the answer is correct.
Returns \texttt{is\_correct}, \texttt{confidence}, and \texttt{reasoning}.
Not used in the main experiments (which all have ground truth).

\section{Data Processing}
\label{app:data}

\subsection{Trace Representation}

All traces are converted to a common \texttt{StructuredTrace} format by the \texttt{TraceExtractor}~\cite{agentops_enabling,beyond_black_box_benchmarking}.
The extractor parses LangGraph conversation state (a JSON file containing the message history, table registry, and agent configuration) into an ordered list of \texttt{Step} objects.
Each step captures:
\begin{itemize}
\itemsep0pt
    \item \texttt{index}: 1-based step position
    \item \texttt{ai\_reasoning}: the agent's chain-of-thought text
    \item \texttt{tool\_name}: tool invoked (e.g., \texttt{f\_filter\_rows})
    \item \texttt{tool\_args}: structured arguments (dict)
    \item \texttt{tool\_result}: tool output string (truncated to 500 chars for prompts)
    \item \texttt{is\_final\_answer}: whether this step is the agent's final response
\end{itemize}

The table context is built from the table registry in the conversation state, listing each table's name, columns, data types, and row count.

\subsection{WikiTableQuestions (WTQ)}

WTQ traces are generated by running the \textsc{Realy} agent (our internal table-QA agent) on the WikiTableQuestions benchmark~\cite{pasupat2015wtq}.
We use 137 failing traces from batch evaluation runs.
Each trace is a LangGraph conversation state JSON file stored at \texttt{data/wikitable\_trails/runs/<id>/conversation\_state.json}.
A results manifest at \texttt{data/wikitable\_trails/results.json} maps test IDs to expected answers and correctness labels.

Human annotations are collected for 119 traces via a review CLI.
Each annotation records the annotator-identified error step index and a free-text error description.
Annotations are stored at \texttt{data/wikitable\_human\_labels/merged.json} with multiple annotators per trace merged into a set of valid error steps.

\subsection{GAIA}

GAIA traces come from the GAIA benchmark~\cite{mialon2023gaia} and use a span-tree structure; they are sourced from the TRAIL benchmark~\cite{trail2025tracelocalization}.
We load 117 failing traces from \texttt{data/trail/manifest.json}.
GAIA conversation states require special handling: spans are converted to LangGraph-compatible message format by mapping each span to an assistant message (with tool calls) and a tool response message.
83 traces have human step-level annotations stored at \texttt{data/trail/human\_annotations.json}.

\subsection{BigBenchMistake (BBM)}

BBM provides chain-of-thought traces with ground-truth \texttt{mistake\_index} labels.
We load 150 traces from \texttt{data/bbm/manifest.json}.
Each trace is a sequence of reasoning steps (not tool calls), so the \texttt{tool\_name} field is set to the step's reasoning type.
The \texttt{mistake\_index} serves as the ground-truth error step for evaluation.
Answer verification uses a task-specific extractor that handles mathematical, multiple-choice, and short-answer formats.

\subsection{SWE-bench}

SWE-bench traces are from software engineering agents with code-execution capabilities.
We load 31 failing traces from \texttt{data/swebench/manifest.json}.
These traces are substantially longer than table-QA traces (median 12 steps vs.\ 4 for WTQ) and include file-editing, test-running, and code-searching actions.
30 traces have human annotations.

\section{Inter-Annotator Agreement}
\label{app:iaa}

The WTQ agent benchmark is collected internally by our team: we run the \textsc{Realy} agent on WikiTableQuestions, collect failing traces, and obtain step-level human annotations from multiple annotators.
Each of the 119 labeled traces was independently annotated by two annotators who identified the earliest decisive error step and wrote a free-text error description.
Annotators were presented with the full agent trace via a review CLI and asked to
(1) identify the index of the earliest step whose correction would enable a correct
final answer, and (2) write a free-text description of the error. No additional
guidelines or examples were provided beyond the task definition.
Pairwise exact-match agreement is 0.78 and Krippendorff's $\alpha$ (ordinal) is 0.82, indicating substantial agreement.
The remaining disagreements reflect genuine ambiguity in identifying the ``first'' error when multiple steps contribute to failure; we retain all annotators' labels as a set-valued gold standard for evaluation.

GAIA and SWE-bench use agent traces and step-level annotations from \citet{trail2025tracelocalization}; BBM uses mistake-index labels from \citet{tyen2024mistakefinding}.
We manually verified that step-level labels align correctly after format conversion on a sample of GAIA and SWE-bench traces.
GAIA and SWE-bench include human annotations from the original benchmark authors.

\section{Error Taxonomy}
\label{app:taxonomy}

The 11-category error taxonomy is derived from 256 SWE-bench human annotations with typo variants merged. The categories are: formatting errors, instruction noncompliance, context handling failures, resource abuse, poor information retrieval, incorrect problem identification, hallucination, tool errors, task orchestration issues, goal deviation, and incorrect memory usage. Full definitions are in the released code.

\section{Reproducibility}
\label{app:reproducibility}

\subsection{Software Environment}

\begin{itemize}
\itemsep0pt
\item Python $\geq$ 3.10
\item \texttt{err-loc} 0.1.0 (custom package; \texttt{pip install -e err\_loc/})
\item \texttt{langchain-openai} $\geq$ 0.2, \texttt{langchain-core} $\geq$ 0.3
\item \texttt{langgraph} $\geq$ 1.0.2, \texttt{langchain} $\geq$ 1.0.3
\item \texttt{pydantic} 2.11.9
\item \texttt{pandas} $\geq$ 2.3.2, \texttt{numpy} $\geq$ 2.1.0
\item \texttt{pytest} 8.4.2
\end{itemize}

\subsection{Model Access}

All experiments use the OpenAI API with model \texttt{gpt-5.2-2025-12-11} at temperature 0.0, unless otherwise noted.
All baselines (AAO, SBS, BS, Reflexion, ICS, DoVer) use the same model and temperature.
The AAO~(Opus~4.6) comparison uses \texttt{claude-opus-4-6} at temperature 0.0.
ThinkPRM uses its published 1.5B checkpoint (local GPU inference). AgentRx uses \texttt{gpt-5.2-2025-12-11} for all LLM calls.
API keys are stored in a local JSON file and are not committed to the repository.

\subsection{Compute}

Experiments were run on a single Linux machine (Ubuntu 22.04, 64 GB RAM).
No GPU is required; all computation is via API calls.
Wall-clock time per benchmark (full \methodname pipeline):
\begin{itemize}
\itemsep0pt
\item WTQ (137 traces): ${\sim}70$ min
\item GAIA (117 traces): ${\sim}60$ min
\item BBM (150 traces): ${\sim}45$ min
\item SWE-bench (31 traces): ${\sim}30$ min
\end{itemize}

The full ablation study (8 configurations $\times$ 137 WTQ traces) takes ${\sim}6$ hours.

\subsection{Evaluation Runner}

All experiments are executed via a unified CLI:

\begin{small}
\begin{verbatim}
python err_loc_demo/scripts/run_eval.py \
  --benchmark wikitable \
  --method reflect_repairer \
  --enable_correction \
  --hard_gate --hard_gate_max_retries 3 \
  --output_dir results/wtq_reflect \
  --auditor_model gpt-5.2-2025-12-11 \
  --agent_model gpt-5.2-2025-12-11
\end{verbatim}
\end{small}

Results are saved incrementally (one JSON entry per completed example), allowing interrupted runs to be resumed.
Each run saves a \texttt{run\_config.json} with all CLI arguments, \texttt{sys.argv}, and a timestamp.
Accuracy reports are computed automatically against human labels when available.

\section{Plan-Anchored Verifier: Design and Evaluation}
\label{app:plan_anchor_verifier}

This section details the plan-anchored verifier introduced in \S\ref{sec:plan_anchor_verifier}, which replaces the oracle for online settings where ground-truth answers are unavailable, and compares it against a composite proxy verifier baseline.

\subsection{Architecture}

The plan-anchored verifier reasons about the coherence between the task specification and the agent's execution via stepwise deviation auditing.
Given a trace~$\tau$ and the original question~$q$, the verifier proceeds in four stages (Figure~\ref{fig:plan_anchor}):

\begin{figure}[t]
\centering
\includegraphics[width=\columnwidth]{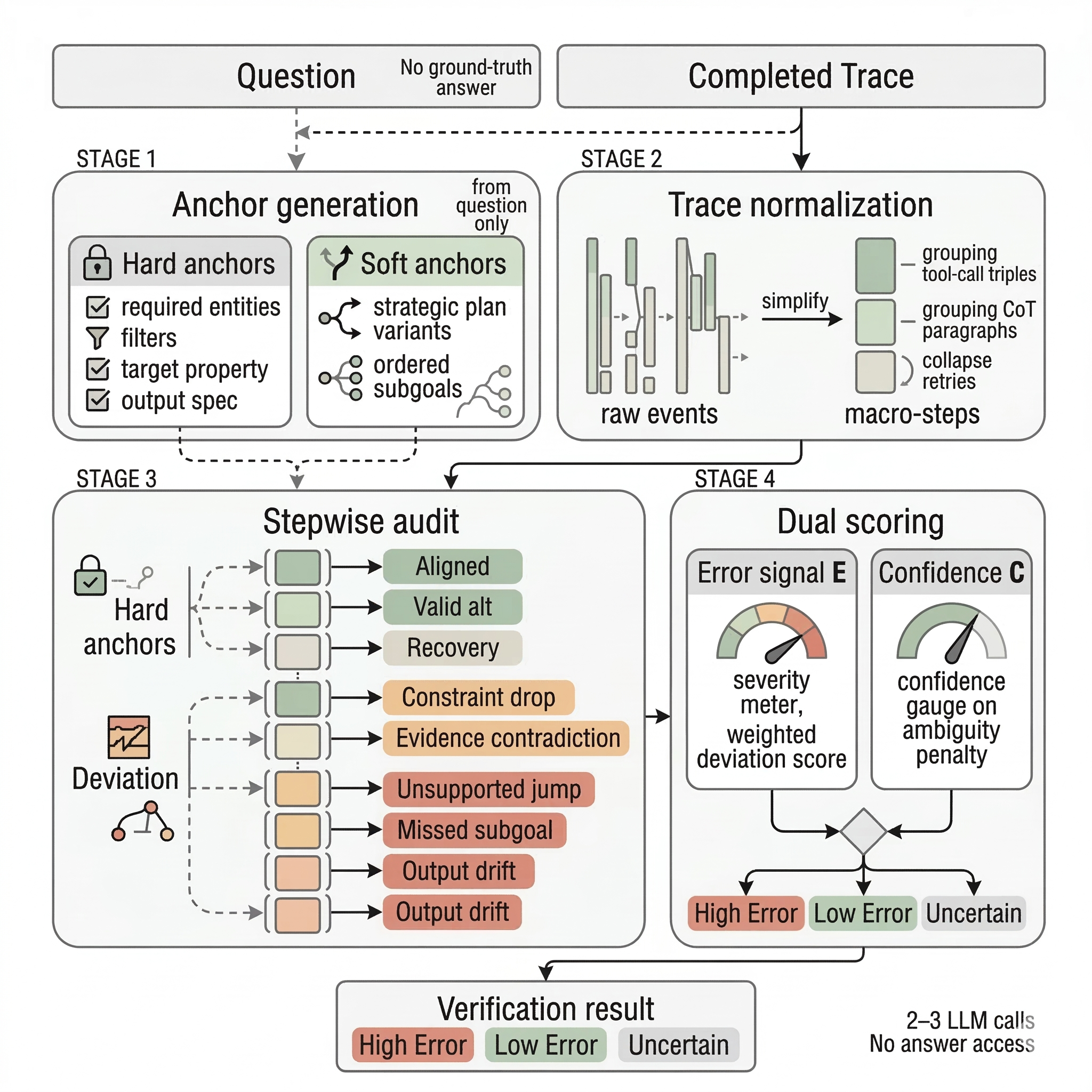}
\caption{Plan-anchored verification as a secondary, no-ground-truth regime: the verifier extracts task anchors, audits normalized macro-steps for deviations, and outputs an error signal with confidence.}
\label{fig:plan_anchor}
\end{figure}

\paragraph{Stage 1: Anchor generation (1 LLM call).}
From the question alone (no ground truth), the verifier synthesizes two types of anchors:
\begin{itemize}
\itemsep0pt
\item \textbf{Hard anchors} (non-negotiable task obligations): required entities/tables, filters and conditions, target property, operation type (count, lookup, compare, aggregate), and output form/spec.
\item \textbf{Soft anchors} (strategic plan variants): 1--2 plausible execution plans, each with ordered subgoals.
\end{itemize}

\paragraph{Stage 2: Trace normalization.}
Trace events are grouped into macro-steps: tool-use traces yield (assistant reasoning, tool call, tool result) triples; chain-of-thought traces are split by paragraph or step markers.
Failed tool calls immediately followed by successful retries are collapsed into single recovery steps.

\paragraph{Stage 3: Stepwise audit (1--2 LLM calls).}
Each macro-step is classified into one of nine categories: four benign (\textsc{Housekeeping}, \textsc{Aligned}, \textsc{Valid\_Alternative}, \textsc{Recovery}) and five deviation types:
\begin{itemize}
\itemsep0pt
\item \textsc{Constraint\_Drop}: a required filter or condition is permanently abandoned.
\item \textsc{Evidence\_Contradiction}: the step contradicts its own tool outputs.
\item \textsc{Unsupported\_Jump}: a conclusion is drawn without supporting evidence.
\item \textsc{Missed\_Required\_Subgoal}: a necessary intermediate step is skipped.
\item \textsc{Output\_Drift}: the final answer is semantically inconsistent with the trace evidence.
\end{itemize}
For long traces (${>}8$ macro-steps), the audit is chunked with 2-step overlap.
Related deviations are merged; deviations followed by \textsc{Recovery} verdicts are marked as resolved.

\paragraph{Stage 4: Dual scoring.}
An error signal $E$ is computed as a severity-weighted sum of deviations (critical/major/minor $\times$ hard/soft, with resolved deviations discounted to 30\%) squashed through a sigmoid to $[0,1]$.
An audit confidence $C$ is penalized for ambiguity (excess \textsc{Valid\_Alternative} steps, many unresolved soft deviations, or too few substantive steps).
The health score is $h = 1 - E$, and the final label is:
\begin{itemize}
\itemsep0pt
\item \textsc{High\_Error\_Signal} if $C \geq 0.3$ and $E \geq 0.5$.
\item \textsc{Low\_Error\_Signal} if $C \geq 0.3$ and $E < 0.5$.
\item \textsc{Uncertain} if $C < 0.3$.
\end{itemize}

\subsection{Evaluation Data: Constructing Correct Traces}
\label{app:correct_traces}

To evaluate verifier discrimination (correct vs.\ incorrect), we construct balanced or near-balanced evaluation sets containing both correct and incorrect traces for each benchmark:

\begin{itemize}
\itemsep0pt
\item \textbf{WTQ ($N{=}502$):} 264 correct + 238 incorrect traces from the same \textsc{Realy} agent batch runs, oracle-verified via string matching.
\item \textbf{BBM ($N{=}300$):} 150 correct (no \texttt{mistake\_index}) + 150 incorrect (with \texttt{mistake\_index}), sampled with a fixed random seed.
\item \textbf{GAIA ($N{=}127$):} 10 correct traces (generated by running the GAIA agent on tasks with oracle-verified correct answers) + 117 failing traces from the TRAIL benchmark.
\item \textbf{SWE-bench ($N{=}45$):} 14 correct traces (passing patches verified by the benchmark's test suite) + 31 failing traces.
\end{itemize}

\noindent WTQ and BBM are balanced by construction.
GAIA (10:117) and SWE-bench (14:31) are genuinely imbalanced; AUROC should be interpreted with caution on these benchmarks, and F1 and error precision are more stable metrics under imbalance.

\subsection{Cross-Benchmark Detection Performance}
\label{app:verifier_eval}

Table~\ref{tab:verifier_comparison_app} reports plan-anchored verifier detection performance.
We report AUROC (health score vs.\ oracle label; higher is better, 0.5 is chance), best error-positive F1 via threshold sweep, and error precision at the best threshold.

\begin{table}[t]
\centering
\small
\caption{Plan-anchored verifier detection performance. AUROC measures discrimination between correct and incorrect traces; 0.5 is chance.}
\label{tab:verifier_comparison_app}
\begin{tabular}{@{}lrrrr@{}}
\toprule
Benchmark & $N$ & AUROC & Best-F1 & Err-Prec \\
\midrule
WTQ & 502 & 0.680 & 0.769 & 0.625 \\
BBM & 300 & 0.561 & 0.702 & 0.568 \\
GAIA & 127 & 0.842 & 0.888 & 0.934 \\
SWE-bench & 45 & 0.968 & 0.933 & 0.966 \\
\bottomrule
\end{tabular}
\end{table}

Discrimination is strongest on GAIA (AUROC~0.84) and SWE-bench (0.97), where the verifier's stepwise deviation auditing effectively distinguishes correct from incorrect multi-step traces.
On WTQ (0.68), the verifier achieves moderate discrimination with strong Best-F1 (0.769).
On BBM (0.56), discrimination is limited: chain-of-thought traces lack the tool-call structure that anchor-based auditing exploits.
The cost is 2--3 LLM calls per trace (anchor generation + stepwise audit).

\section{Benchmark-Specific Agent Configurations}
\label{app:agent_configs}

\subsection{WTQ Agent (\textsc{Realy})}

The \textsc{Realy} agent is a LangGraph-based table-QA system with 14 tabular-algebra tools. Traces are stored as conversation state JSON files.

\subsection{GAIA Agent}

The GAIA agent has access to web search, page navigation, file search, text analysis, and answer submission tools. Traces are converted from TRAIL's span-tree format to our common representation by mapping each span to an assistant message with tool calls and a tool response message.

\subsection{BBM (BigBenchMistake)}

BBM traces contain chain-of-thought reasoning without tool calls.
Each reasoning step is treated as a separate localization unit with \texttt{tool\_name} set to the step's reasoning type.
\methodname applies its full diagnose-replay-verify pipeline on BBM, including re-execution of the chain-of-thought reasoning.
Evaluation uses pre-existing traces with ground-truth \texttt{mistake\_index} labels.

\subsection{SWE-bench Agent}

The SWE-bench agent uses a \texttt{CodeAgent} (smolagents) that executes Python code directly at each step rather than calling named tools. Each step comprises LLM-generated code for file reading/editing, searching, and test execution via Python standard library and subprocess calls. The only explicit tool is \texttt{FinalAnswerTool}.

\section{Computational Cost}
\label{app:cost}

Table~\ref{tab:cost} reports the theoretical call complexity and empirical token usage per trace for each method.
We measure tokens consumed during actual evaluation runs (all using \texttt{gpt-5.2}) to show that \methodname's accuracy gains are \emph{not} attributable to higher compute budgets.

\begin{table*}[t]
\centering
\small
\caption{Computational cost per trace. \textbf{Left:} theoretical LLM call complexity ($T$ = trace steps, $D$ = max repair iterations, $K$ = rollback points, $N$ = AgentRx constraint checks where $N{\geq}T$). \textbf{Right:} empirical average tokens per trace on \texttt{gpt-5.2-2025-12-11}. Values with ${\sim}$ are trace-length-scaled estimates from measured total token costs; all others are measured.}
\label{tab:cost}
\begin{tabular}{@{}lc rrrr@{}}
\toprule
 & Calls/trace & \multicolumn{4}{c}{Empirical tokens/trace} \\
\cmidrule(l){3-6}
Method & (theoretical) & WTQ & GAIA & SWE & BBM \\
\midrule
\multicolumn{6}{@{}l}{\emph{Localization-only baselines}} \\[2pt]
AAO & 1 & 985 & ${\sim}$2{,}500 & ${\sim}$15{,}000 & ${\sim}$1{,}400 \\
SBS & $T$ & 1{,}854 & ${\sim}$8{,}000 & ${\sim}$75{,}000 & ${\sim}$5{,}500 \\
BS & $\lceil\log_2 T\rceil{+}1$ & 1{,}778 & ${\sim}$3{,}500 & ${\sim}$18{,}000 & ${\sim}$2{,}500 \\
Reflexion & 2 & ${\sim}$1{,}100 & ${\sim}$2{,}800 & ${\sim}$16{,}000 & ${\sim}$1{,}500 \\
DoVer & $N{+}2$ & 3{,}021 & 4{,}715 & 30{,}464 & 4{,}157 \\
ThinkPRM & 0 (GPU) & 0 & 0 & 0 & 0 \\
AgentRx & $N{+}2$ & \textbf{20{,}023} & \textbf{47{,}068} & \textbf{157{,}599} & \textbf{31{,}654} \\
\midrule
\multicolumn{6}{@{}l}{\emph{Correction methods}} \\[2pt]
ICS & ${\leq}D{\times}2$ & 6{,}996 & ${\sim}$15{,}000 & 30{,}214 & ${\sim}$9{,}000 \\
Reflexion{+}ICS & ${\leq}D{\times}2{+}2$ & 11{,}786 & 9{,}812 & 38{,}977 & ${\sim}$11{,}000 \\
\methodname & ${\leq}D{\times}K{\times}3$ & ${\sim}$12{,}000 & ${\sim}$28{,}000 & 53{,}724 & ${\sim}$18{,}000 \\
\bottomrule
\end{tabular}
\end{table*}

\paragraph{Key observations.}
(1)~\textbf{AgentRx is the most compute-intensive method by a wide margin}, consuming 20{,}023--157{,}599 tokens per trace, 3--20$\times$ more than localization-only baselines and 1.5--3$\times$ more than \methodname on SWE-bench.
Despite this, AgentRx achieves 51.1\% localization EM on WTQ (vs.\ \methodname's 76.3\%) and produces no correction output.
On SWE-bench, AgentRx uses 157{,}599 tokens/trace (${\sim}$3$\times$ \methodname's 53{,}724) yet achieves only 41.9\% EM.
(2)~\textbf{ICS is cheaper per trace than \methodname} (7{,}K vs.\ 12{,}K on WTQ, 30{,}K vs.\ 54{,}K on SWE-bench), but achieves substantially lower correction rates (23.7\% on WTQ, 12.0\% on GAIA, and 85.2\% on SWE-bench, versus \methodname's 92.6\%, 54.2\%, and 89.0\%).
The additional compute \methodname spends on diagnosis, gating, and multi-rollback produces qualitatively different outcomes.
(3)~Reflexion{+}ICS (Table~2) provides a direct coupling control: it uses comparable compute to ICS (6.8--6.7 calls/trace) with a different localizer (Reflexion instead of all-at-once), and achieves 0--7.7\% correction, confirming that \methodname's gains come from its diagnostic pipeline, not from additional compute.
Wall-clock time per benchmark is reported in \S\ref{app:reproducibility}.

\section{Faithfulness Details}
\label{app:faithfulness}

This section provides the remaining faithfulness tests (C--E) and the plan-quality analysis omitted from the main paper (\S\ref{sec:faithfulness}).

\paragraph{Test C: Robustness.}
The effect is robust across wording, position, and formatting variants: paraphrase (80.7\%), standard correct (78.7\%), prefix-positioned (77.0\%), and plain-text (74.7\%) all perform within 6pp of each other (Table~\ref{tab:faithfulness}).

\paragraph{Test D: Probability shift.}
The correct hint increases the probability of the target tool by $+17.3$pp (CI: $[+8.0, +28.7]$) at the first resampled step, with a log-odds shift of $+4.95$.

\paragraph{Test E: First divergence.}
In 96.7\% of cases, the hint changes agent behavior at the very first regenerated step (average first divergence: step 3.09), confirming that repair guidance acts immediately.

\paragraph{Does faithfulness help accuracy?}
When the repair plan correctly identifies the error area (step distance $\leq 3$) \emph{and} the agent follows it, correction accuracy reaches 88\%, compared to 75\% when the plan is good but not followed ($+13$pp).
For poor plans (distance $> 3$), adherence provides only $+2$pp (73\% vs.\ 71\%).
This confirms the causal prediction: faithfulness helps most when the plan is worth following.

\section{Explanation Quality}
\label{app:explanations}

Table~\ref{tab:explanation} reports explanation quality on the three benchmarks with human error descriptions (WTQ, GAIA, SWE-bench).
Explanation quality is step-conditioned: only scored when the predicted step matches a human-annotated step.
The number of step-matched entries varies by method and benchmark.

\begin{table*}[t]
\centering
\small
\caption{Error explanation quality on benchmarks with human error descriptions. \emph{Step-cond.}: mean LLM-judge semantic similarity (0--1) scored only on step-matched entries (std: sample standard deviation of similarity scores). High = fraction with similarity $\geq 0.7$ (std: SE of proportion). BBM excluded (no human descriptions). Step-matched $n$ (WTQ\,/\,GAIA\,/\,SWE): AAO 66/13/10, SBS 60/1/8, BS 64/11/10, Reflexion+GT 63/13/14, Reflexion--noGT 63/15/13, AgentRx 61/6/13, ICS 58/9/5, \methodname\ 91/33/21, \methodname\ (proxy) 74/28/20.}
\label{tab:explanation}
\begin{tabular}{@{}l cc cc cc@{}}
\toprule
& \multicolumn{2}{c}{WTQ} & \multicolumn{2}{c}{GAIA} & \multicolumn{2}{c}{SWE-bench} \\
\cmidrule(lr){2-3} \cmidrule(lr){4-5} \cmidrule(lr){6-7}
Method & Sim & High\% & Sim & High\% & Sim & High\% \\
\midrule
AAO & 0.320\tss{.18} & 18.2\tss{4.7} & 0.385\tss{.20} & 30.8\tss{12.8} & 0.150\tss{.12} & 0.0 \\
SBS & 0.285\tss{.17} & 13.3\tss{4.4} & 0.350\tss{--} & 0.0 & 0.269\tss{.16} & 25.0\tss{15.3} \\
BS & 0.305\tss{.18} & 15.6\tss{4.5} & 0.277\tss{.18} & 18.2\tss{11.6} & 0.210\tss{.14} & 0.0 \\
Reflexion+GT & 0.655\tss{.16} & 69.4\tss{5.4} & 0.537\tss{.19} & 47.4\tss{11.5} & 0.196\tss{.13} & 7.1\tss{6.9} \\
Reflexion--noGT & 0.672\tss{.15} & 72.6\tss{5.2} & 0.547\tss{.19} & 42.9\tss{10.8} & 0.265\tss{.16} & 23.1\tss{11.7} \\
ThinkPRM & -- & -- & -- & -- & -- & -- \\
AgentRx & 0.601\tss{.17} & 60.0\tss{5.9} & 0.436\tss{.20} & 37.5\tss{17.1} & 0.389\tss{.19} & 30.8\tss{12.8} \\
ICS & 0.250\tss{.16} & 10.3\tss{4.0} & 0.230\tss{.16} & 11.1\tss{10.5} & 0.125\tss{.10} & 0.0 \\
\midrule
\textbf{\methodname} & \textbf{0.760}\tss{.18} & \textbf{70.0}\tss{5.3} & \textbf{0.610}\tss{.20} & \textbf{53.8}\tss{12.4} & \textbf{0.450}\tss{.19} & \textbf{42.9}\tss{10.8} \\
\methodname (proxy) & 0.640\tss{.18} & 67.9\tss{5.9} & 0.635\tss{.19} & 66.2\tss{9.8} & 0.430\tss{.19} & 38.5\tss{13.5} \\
\bottomrule
\end{tabular}
\end{table*}

On SWE-bench, \methodname (0.450) leads all methods, including AgentRx (0.389), with comparable sample sizes, indicating that the advantage reflects genuine explanation quality rather than selection bias from small $n$.

Among the newly evaluated baselines, Reflexion variants achieve moderate step-conditioned similarity on WTQ (Reflexion--noGT: 0.672; Reflexion+GT: 0.655) and GAIA (Reflexion--noGT: 0.547; Reflexion+GT: 0.537), but remain below \methodname's scores on both benchmarks.
AgentRx also produces strong explanations (WTQ: 0.601).
However, \methodname's higher EM means its explanation scores are computed over a larger and more representative sample of errors.
On SWE-bench, where traces are longer and more complex, \methodname's diagnosis-driven approach yields the best explanations (0.450 vs.\ 0.389 for AgentRx), while Reflexion degrades sharply (0.196--0.265).
ThinkPRM produces only mechanical score listings without explanatory text and is marked as~``--''.

The proxy pipeline achieves comparable step-conditioned scores (0.640 on WTQ, 0.635 on GAIA, 0.430 on SWE-bench), confirming that when proxy localization identifies the correct step, the explanation mechanism produces similar-quality attributions regardless of the verifier regime.

The key advantage of \methodname's explanations emerges through the correction--explanation coupling (\S\ref{sec:correction_coupling}): when correction succeeds, post-correction re-localization produces root-cause explanations grounded in contrastive evidence, a mechanism unavailable to localization-only baselines.

\paragraph{Coverage-aware explanation quality.}
Table~\ref{tab:explanation} reports quality only for step-matched cases, so methods with lower EM benefit from a selection effect.
To account for coverage differences, we compute a coverage-aware metric that assigns similarity~0 to step-unmatched predictions: $\text{Sim}_{\text{cov}} = \text{Sim} \times (n_{\text{matched}} / n_{\text{total}})$.
On WTQ ($n_{\text{total}}=119$): \methodname $0.760 \times (91/119) = 0.581$; AAO $0.320 \times (66/119) = 0.177$; ICS $0.250 \times (58/119) = 0.122$.
On GAIA ($n_{\text{total}}=83$): \methodname $0.610 \times (33/83) = 0.242$; AAO $0.385 \times (13/83) = 0.060$; Reflexion+GT $0.537 \times (13/83) = 0.084$.
On SWE-bench ($n_{\text{total}}=30$): \methodname $0.450 \times (21/30) = 0.315$; AgentRx $0.389 \times (13/30) = 0.169$; Reflexion+GT $0.196 \times (14/30) = 0.091$.
Under this metric, \methodname's advantage increases because its higher localization accuracy means a larger fraction of predictions receive non-zero explanation scores.

\section{Verified vs.\ Fallback Localization}
\label{app:verified_fallback}

Table~\ref{tab:verified_fallback_app} breaks down \methodname's localization accuracy by correction outcome: \emph{verified} traces are those where the repair loop produced a correct answer (the intervention succeeded), and \emph{fallback} traces are those where correction failed and the localizer's best prediction is used without intervention support.

\begin{table}[t]
\centering
\small
\caption{Localization accuracy stratified by correction outcome. Coverage = correction rate from Table~\ref{tab:correction}. Verified EM reflects intervention-supported attribution; Fallback EM reflects localization without intervention evidence.}
\label{tab:verified_fallback_app}
\begin{tabular}{@{}lrcccc@{}}
\toprule
Benchmark & Cov.\% & \multicolumn{2}{c}{Verified} & \multicolumn{2}{c}{Fallback} \\
\cmidrule(lr){3-4} \cmidrule(lr){5-6}
& & EM & Off-1 & EM & Off-1 \\
\midrule
WTQ & 92.6 & 74.1 & 80.6 & 100.0 & 100.0 \\
GAIA & 54.2 & 47.1 & 72.5 & 25.8 & 74.2 \\
BBM & 95.0 & 35.8 & 53.7 & 14.3 & 42.9 \\
SWE-bench & 89.0 & 77.7 & 75.0 & 0.0 & 100.0 \\
\bottomrule
\end{tabular}
\end{table}

On GAIA (47.1\% vs.\ 25.8\% EM) and BBM (35.8\% vs.\ 14.3\%), verified cases substantially outperform fallback, confirming that intervention-supported attribution produces stronger localization than pattern-matching alone.
On SWE-bench (77.7\% vs.\ 0.0\%), verified cases dominate, though the fallback partition ($N{=}3$) is too small for reliable comparison.
On WTQ, where correction coverage is 92.6\%, only 10 traces fall back, and those few happen to be correctly localized (100\% EM), consistent with WTQ being the easiest benchmark where the initial localizer is already strong.
Overall, \methodname provides calibrated output: users can distinguish which attributions are intervention-backed and which are tentative.

\section{Case Study}
\label{app:case_study}

Figure~\ref{fig:case_study} contrasts \methodname with ICS on a representative WTQ example.

\begin{figure}[t]
\centering
\footnotesize
\fcolorbox{black}{gray!10}{\parbox{0.92\columnwidth}{%
\textbf{WTQ example} \quad \emph{Q: ``Which country has the most cities with population over 500k?''}\\[2pt]
Steps 1--3: load table, inspect columns, sort by population. \textbf{Step~4}: \texttt{f\_filter\_rows(table=t1, column="population", condition=">500000")} $\to$ returns 12 rows (correct).\\
\textbf{Step~5}: \texttt{f\_group\_by(table=t1, column="country")} $\to$ groups by \emph{unfiltered} table (silent bug: should use \texttt{t1\_filtered}). Agent answers ``China''\ (wrong; gold = ``United States'').
}}

\vspace{4pt}

\fcolorbox{darkgreen}{green!8}{\parbox{0.92\columnwidth}{%
\textbf{\ding{51}\ \methodname} \quad Localizes step~4 (filter), diagnoses stale-reference error.\\
Rolls back to step~4, injects: \emph{``pass filtered output to group-by.''}\\
Replayed agent answers ``United States''\ (\textbf{correct}). Post-correction re-localization confirms step~4 with similarity 0.95.
}}

\vspace{4pt}

\fcolorbox{red}{red!8}{\parbox{0.92\columnwidth}{%
\textbf{\ding{55}\ ICS} \quad Resamples from step~5 prefix without guidance.\\
3 iterations: agent re-derives same group-by on unfiltered table each time.\\
Final answer ``China''\ (\textbf{wrong}); explanation: ``the agent should have been more careful with grouping.''
}}

\caption{Case study on a WTQ example. \methodname produces a step-specific, outcome-grounded attribution tied to its successful intervention; ICS retries without targeting the root cause and produces a vague explanation.}
\label{fig:case_study}
\end{figure}

In the \methodname trace, the localizer flags step~4 (the filter whose output is silently dropped); after rollback and guided replay, the agent passes the filtered table to group-by and answers correctly. Post-correction re-localization compares the two continuations and confirms step~4 with similarity 0.95, pinpointing the stale-reference bug.
ICS, lacking targeted guidance, resamples from the step-5 prefix three times; each retry re-derives the same group-by on the unfiltered table and returns the wrong answer, with an explanation (``the agent should have been more careful with grouping'') that names no step and no mechanism, illustrating the correction--explanation coupling gap quantified in Table~\ref{tab:correction}.

\section{Extended Positioning Against Baselines}
\label{sec:extended_related_work}

This section provides detailed positioning of \methodname against each baseline along the four attribution requirements identified in \S\ref{sec:intro}: execution grounding~(R1), prefix-preserving replay~(R2), targeted intervention~(R3), and inference-time computation~(R4).
Each subsection describes what the baseline does, which requirements it satisfies, and what specific gap \methodname addresses.
The assignments below correspond to Table~\ref{tab:method_comparison}.

\subsection*{Reflexion \cite{shinn2023reflexion}}

Reflexion generates a verbal self-critique after a failed attempt and retries with the critique appended to the prompt.
Its primary goal is correction (recovering a correct answer), not step-level attribution.

\textbf{R1} (\ding{55}): The ``explanation'' is a free-text self-reflection generated by the same model, not grounded in execution evidence.
Such self-critiques are known to be vulnerable to post-hoc rationalization~\cite{turpin2023cotunfaithful,huang2023selfcorrect,tyen2024mistakefinding}.
\textbf{R2} (\ding{55}): The retry generates an independent trajectory; there is no prefix-preserving replay, so even a successful retry does not identify which step in the \emph{original} trace was decisive.
\textbf{R3} (\ding{55}): The critique is holistic (e.g., ``try a different approach'') rather than a targeted intervention at a specific step.
\textbf{R4} (\checkmark): Operates at inference time.

\methodname's correction--localization coupling (Table~\ref{tab:correction}) directly quantifies this gap: Reflexion's $\Delta$ between corrected and failed explanation quality is $\leq +0.05$, while \methodname's is $+0.25$ to $+0.29$, confirming that \methodname's corrections are coupled to understanding while Reflexion's are not.

\subsection*{ICS \cite{ics2025backtrack}}

ICS structures reasoning into discrete thought steps, localizes an error via self-inspection, backtracks to the localized step, and resamples from the bare prefix without any repair guidance.
Its goal is correction via structured backtracking.

\textbf{R1} (\halfmark): ICS does execute replay, but does not use the outcome to ground or refine the attribution; success or failure of the resample is not fed back to sharpen the localization.
\textbf{R2} (\checkmark): Resamples from a prefix of the original trace.
\textbf{R3} (\ding{55}): This is the critical gap. ICS resamples from a bare prefix without any repair plan, correction instruction, or constrained guidance.
A success may reflect stochastic variation rather than identification of the error.
\methodname's faithfulness ablation (Table~\ref{tab:faithfulness}) directly tests this: the no-hint baseline achieves 35.6\% accuracy vs.\ 78.7\% for the correct hint, demonstrating that bare resampling is a weak test of error identification.
\textbf{R4} (\checkmark): Operates at inference time.

An additional structural difference is that ICS requires traces to be generated in its Thought MDP format from the start, whereas \methodname works post-hoc on any agent trace without requiring control over the generation process.
\methodname's targeted intervention (diagnosis $\to$ repair plan $\to$ constrained replay) and post-correction re-localization close the loop that ICS leaves open.

\subsection*{Trained Classifiers and Constraint Inspectors}

AgenTracer~\cite{agentracer2025localization}, TrajAD~\cite{liu2026trajad}, and ThinkPRM~\cite{thinkprm2025,choudhury2025agentprm,xi2025agentprm} produce step-level predictions at inference time without executing any intervention, violating R1--R3. AgenTracer uses counterfactual replay during data curation but distills the signal into a fixed 8B model; TrajAD trains on synthetically perturbed trajectories; ThinkPRM scores steps via a learned reward model. All three are distribution-bound: their predictions reflect training patterns rather than test-time execution evidence (R4 violated for AgenTracer/TrajAD; satisfied for ThinkPRM via GPU inference).

AgentRx~\cite{agentrx2026} synthesizes constraints from tool schemas and produces structured validation logs (R4 satisfied), but never tests whether fixing a flagged step changes the outcome (R1--R3 violated).

\subsection*{DoVer \cite{ma2025dover}}

DoVer segments a trace into trials at re-plan boundaries, then applies a four-stage pipeline: (1)~trial segmentation, (2)~failure attribution via an LLM summarizer that hypothesizes a faulty step or agent per trial, (3)~intervention generation that turns the hypothesis into a concrete edit, and (4)~intervention execution with differential evaluation using task success and milestone progress.

The architectural distinction from \methodname is in the information flow. DoVer's localization (Stage~2) is produced by an LLM log summarizer, functionally equivalent to an all-at-once judge, that operates entirely on the log without execution-based evidence. The intervention in Stage~4 validates whether the hypothesized fix resolves the failure, classifying the outcome as Validated, Partially Validated, Refuted, or Inconclusive. Crucially, this classification is not fed back to revise the localized step: if an intervention fails, the hypothesis is labeled Refuted, but no re-localization occurs. The replay tests ``does this fix help?'' not ``is this step the causal origin?''

\textbf{R1} (\halfmark): DoVer executes interventions and observes outcomes, which constitutes partial execution grounding. However, the outcome is used to validate the debugging hypothesis, not to refine step-level attribution.
\textbf{R2} (\halfmark): DoVer replays from the intervention point preserving the prefix, but does not use the corrected continuation as contrastive evidence to sharpen attribution.
\textbf{R3} (\checkmark): Interventions are targeted and diagnosis-specific.
\textbf{R4} (\checkmark): Operates at inference time.

DoVer explicitly argues that single-step attribution is often ill-posed and shifts focus to outcome-oriented progress metrics, measuring whether the system makes quantifiable progress toward task success rather than evaluating localization accuracy. \methodname takes the opposite position: it demonstrates that precise step-level attribution \emph{is} achievable when correction is used as contrastive evidence. Post-correction re-localization, the component that closes the loop from correction back to attribution, has no analogue in DoVer's pipeline. The ablation in Table~\ref{tab:ablation} ($+10.6$pp EM from re-localization alone) quantifies the value of this closed loop.

In our empirical evaluation (Table~\ref{tab:localization}), DoVer's localization performance is comparable to the AAO baseline, consistent with its use of an equivalent LLM summarizer for failure attribution.

TRAIL~\cite{trail2025tracelocalization} is an evaluation framework (not a localization method) already discussed in \S\ref{sec:related}. Our prompt-based baselines (AAO, SBS, BS) are described in Appendix~\ref{app:baselines}.

\section{Dataset Licenses and Terms of Use}
\label{app:licenses}
 
Our evaluation uses four benchmarks, each derived from publicly available resources:
 
\begin{itemize}
\item \textbf{WTQ traces.} We run a tabular agent on a subset of WikiTableQuestions \citep{pasupat2015wtq}, which is released under the \href{https://creativecommons.org/licenses/by/4.0/}{CC-BY-4.0} license. Our contribution is the 137 failing agent traces and 119 step-level human annotations collected internally; these are included in the supplementary material and will be released under CC-BY-4.0 upon acceptance.
 
\item \textbf{GAIA and SWE-bench traces.} We use subsets of the TRAIL benchmark \citep{trail2025tracelocalization}, which provides multi-step agent traces with human step-level annotations over GAIA \citep{mialon2023gaia} and SWE-bench \citep{jimenez2023swebench} tasks. GAIA is released under CC-BY-4.0; SWE-bench is released under the MIT license. The TRAIL traces and annotations are used in accordance with their publicly released terms.
 
\item \textbf{BBM.} BigBenchMistake \citep{tyen2024mistakefinding} provides chain-of-thought traces with ground-truth mistake indices. BigBench is released under the Apache-2.0 license.
\end{itemize}
 
All benchmark data are used for research evaluation consistent with the original intended use. The \texttt{err-loc} library (pipeline code, baselines, evaluation scripts) included in the supplementary material will be released under the MIT license upon acceptance.
\section{Limitations}
\label{sec:limitations}

This work has several limitations.
First, \methodname is evaluated on relatively structured silent-failure settings (table-QA, multi-hop reasoning, chain-of-thought, and software engineering traces); extending reliable intervention-based localization to more unstructured domains remains an open challenge.
Second, the primary evaluation regime relies on oracle access to the expected answer for verification; while the plan-anchored proxy verifier provides a deployment-time alternative, it is weaker than oracle-verified attribution and should not be treated as equivalent evidence.
Third, replay-based attribution requires re-executing the agent, which may not be feasible in all environments (e.g., when tool calls have irreversible side effects or when the execution environment is no longer available).
Fourth, the method localizes a single earliest decisive error step; traces with multiple independent errors or distributed failure modes may require multi-point attribution beyond the current framework.

\section{Ethical Considerations}

\methodname is designed for post-hoc debugging and auditing of LLM agent traces, and we do not foresee direct negative societal impacts from this work.
The method operates on completed traces and does not modify deployed agent behavior in real time.
All benchmarks used in this paper are publicly available or internally collected with appropriate oversight; no personal or sensitive data is involved.
We note that automated error attribution could be misused to assign blame in high-stakes settings without adequate human oversight~\cite{nistairmf,euaiact12}; we recommend that \methodname's output be treated as a diagnostic aid rather than a definitive judgment, and that human review remain part of any deployment pipeline.

The \texttt{err-loc} library (core pipeline, all baselines, evaluation metrics) and internally generated WTQ agent traces with human step-level error annotations are included in the supplementary material.

\end{document}